\documentclass[10pt,twocolumn,letterpaper]{article}
\usepackage[T1]{fontenc}
\usepackage{cvpr}
\usepackage{pifont}
\usepackage{multirow}
\usepackage{makecell}
\usepackage{array}
\usepackage{colortbl}
\usepackage{placeins}
\newsavebox{\ablA}\newsavebox{\ablB}\newsavebox{\ablC}\newsavebox{\ablD}
\newlength{\ablL}\newlength{\ablR}
\newcommand{\subtab}[2]{\parbox[t]{\wd#1}{\usebox{#1}\par\vspace{4pt}\raggedright #2}}

\newcommand{\tactus}{TACTUS\xspace}
\newcommand{\net}{TactusNet\xspace}
\newcommand{\step}[1]{\ding{\numexpr171+#1\relax}}
\newcommand{\yes}{$\bullet$}
\newcommand{\no}{$\circ$}

\setlist[itemize]{leftmargin=*,topsep=2pt,itemsep=1pt,parsep=0pt}
\makeatletter
\renewcommand\paragraph{\@startsection{paragraph}{4}{\z@}{1.2ex \@plus .3ex \@minus .2ex}{-1em}{\normalfont\normalsize\bfseries}}
\makeatother

\definecolor{cvprblue}{rgb}{0.21,0.49,0.74}
\usepackage[breaklinks,colorlinks,allcolors=cvprblue]{hyperref}
\hypersetup{hypertexnames=false,pdftitle={TACTUS: Unifying Surgical Action Understanding in Time and Space},pdfauthor={Lurui Wang, Hao Ding, Chenhao Yu, Chenyan Jing, Jiru Xu, Ruoqi Zuo, Liwei Mei, Chang Liu, Lalithkumar Seenivasan, Mathias Unberath}}
\title{TACTUS: Unifying Surgical Action Understanding in Time and Space}
\author{Lurui Wang\textsuperscript{1,*}\quad Hao Ding\textsuperscript{1,*}\quad Chenhao Yu\textsuperscript{2}\quad Chenyan Jing\textsuperscript{1}\quad Jiru Xu\textsuperscript{1}\\
Ruoqi Zuo\textsuperscript{1}\quad Liwei Mei\textsuperscript{1}\quad Chang Liu\textsuperscript{3}\quad Lalithkumar Seenivasan\textsuperscript{1}\quad Mathias Unberath\textsuperscript{1,\ensuremath{\dagger}}\\[.45em]
{\normalsize\textsuperscript{1}Johns Hopkins University \qquad
\textsuperscript{2}University of Virginia, Charlottesville \qquad
\textsuperscript{3}University of Toronto}\\[.25em]
{\small \textsuperscript{*}Equal contribution \enspace \href{mailto:lwang255@jhu.edu}{lwang255@jhu.edu}, \href{mailto:hding2455@gmail.com}{hding2455@gmail.com} \quad
\textsuperscript{\ensuremath{\dagger}}Corresponding author \enspace \href{mailto:mathias@jhu.edu}{mathias@jhu.edu}}}

\begin{document}
\maketitle
\begingroup
\captionsetup{skip=4pt}
\begin{figure*}[t]
  \centering
  \includegraphics[width=0.86\textwidth]{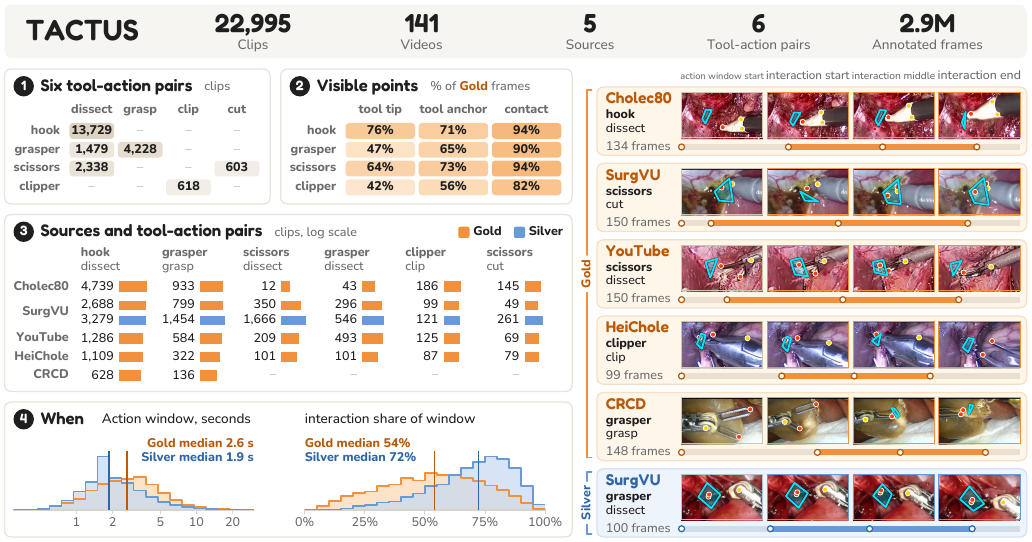}
  \caption{\textbf{Overview of \tactus.} The 22,995 clips cover six tool--action pairs; each pair includes Gold clips from at least four sources.
  The top panels show pair counts, Gold point visibility, source--pair counts (log scale), and tier-normalized temporal statistics.
  On the right, one example per pair shows the action-window start and the interaction-window start, midpoint, and end. Bars mark interaction windows.
  Orange: Gold. Blue: Silver.}
  \label{fig:overview}
\end{figure*}
\makeatletter
\renewcommand\paragraph{\@startsection{paragraph}{4}{\z@}{1.1ex \@plus .3ex \@minus .2ex}{-1em}{\normalfont\normalsize\bfseries}}
\makeatother
\begin{abstract}
Advancing surgical automation requires a comprehensive spatio-temporal understanding of how instruments act on tissue throughout an action.
Existing surgical benchmarks, however, largely define supervision through task-specific targets, leaving the relationships between action semantics, instrument motion, and tissue interaction incompletely specified.
To address this limitation, we introduce \mbox{TACTUS} (Tool, Action, Contact and Trajectory Understanding in Surgery), a new dataset and benchmark that establishes a unified annotation framework for complete surgical actions.
The framework jointly specifies tool and action identities, nested action and interaction windows, instrument-specific landmark trajectories, and tissue-region tracks.
To construct these annotations at scale, we develop a two-tier pipeline combining human annotation, point tracking, and model-generated spatial prompts.
Using this pipeline, we re-annotate 141 videos from five public sources, producing 22,995 action clips spanning 2.9 million frames.
The dataset comprises 15,668 Gold clips with human spatial prompts followed by review and re-checking, and 7,327 Silver clips combining human semantic and temporal annotations with model-generated spatial prompts.
We further introduce \net, the companion reference network, to benchmark semantic, temporal, and spatial prediction from shared video features.
Across \net and 13 published baselines, tool--action accuracy reaches 95.8\% with per-head encoder fine-tuning, while the best temporal recall and contact Dice reach 46.6\% at temporal IoU 0.7 and 0.351, respectively.
Moreover, adding Silver improves the default \net's pair accuracy from 93.8\% to 95.5\%.
Together, these contributions establish a foundation for learning and evaluating the instrument--tissue interactions underlying future surgical automation.
\newline Project page: {\urlstyle{same}\url{https://ruifreddie.github.io/TACTUS/}}.
\end{abstract}

\section{Introduction}
\label{sec:intro}

Advancing surgical automation requires a comprehensive spatio-temporal understanding of how instruments act on tissue.
Such understanding requires supervision that links which instrument acts (Which), what it does (What), when it acts (When), how it moves (How), and where it interacts with tissue (Where), collectively termed 4W1H.
Their annotations must therefore preserve a common action identity from approach through execution to completion.

Existing surgical datasets provide valuable supervision for action recognition, temporal localization, and instrument tracking~\cite{nwoye2022rendezvous,chen2026prostatd,nwoye2025cholectrack20,rueckert2026video,wu2025surgpose}.
However, their annotation targets often describe selected aspects of an interaction, leaving its full temporal and spatial extent unspecified.
Temporally, an interval restricted to active execution excludes the surrounding approach and completion, leaving instrument behavior during these stages outside the action annotation.
For spatial supervision, SurgAM expands a single annotated point in a pre-contact frame into an affordance mask~\cite{song2026surgam}.
However, SurgAM's mask describes the actionable tissue region only in the pre-contact frame, leaving the region's subsequent movement and deformation unannotated.
Consequently, execution-only intervals omit the temporal context surrounding the maneuver, while SurgAM's pre-contact masks omit the evolution of tissue involvement during execution.
These gaps leave the relationship between instrument motion and tissue involvement incompletely specified throughout a complete surgical action.

\begin{figure*}[t]
  \centering
  \includegraphics[width=0.94\textwidth]{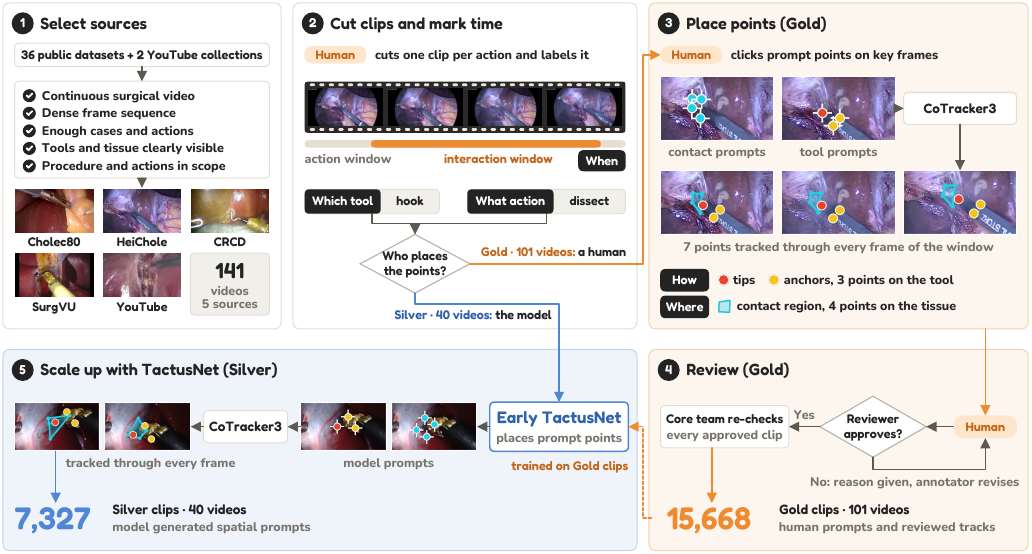}
  \caption{\textbf{Construction pipeline.} Humans label the classes and times of every clip. Gold then adds human prompt points, review, and a full re-check, whereas Silver hands only the points to a model.
  \step{1}--\step{2}~Source selection, clip extraction, and temporal annotation. \step{3}--\step{4}~Gold points and review. \step{5}~Silver points from an early version of \net.
  Orange: Gold. Blue: Silver. The dashed arrow denotes model training.}
  \label{fig:pipeline}
\end{figure*}

Constructing such joint supervision at scale introduces a further challenge.
Instrument landmarks and tissue regions must be followed throughout each action despite occlusion, deformation, and camera motion.
Maintaining these correspondences requires careful trajectory review.
Repeating prompt placement, track inspection, and error correction for additional clips creates a growing manual workload.
A scalable process must therefore preserve supervision quality while reducing the effort required to annotate additional clips.

\paragraph{Contributions.}
To address these annotation and construction challenges, we establish a new protocol with an individual action or attempt as its fundamental unit.
Within this unit, the protocol jointly defines tool and action identities, a complete action window and an inner interaction window, instrument-specific tip and anchor trajectories, tissue-region tracks, and point visibility.

To enable larger-scale annotation with limited human effort, we develop a two-tier pipeline that combines human-reviewed supervision with model-assisted expansion (\cref{fig:pipeline}).
Gold uses human spatial prompts, trajectory propagation, and review to produce checked spatial annotations~\cite{karaev2025cotracker3}.
These annotations also support training a model to generate spatial prompts for additional clips.
\mbox{Building on} this capability, Silver expands the collection through model-generated spatial prompts while retaining human tool, action, and temporal labels.

Using this pipeline, we construct \href{https://ruifreddie.github.io/TACTUS/}{\mbox{TACTUS}} (Tool, Action, Contact and Trajectory Understanding in Surgery), a large-scale dataset and benchmark for following surgical actions in time and space.
We re-annotate 141 videos from five public sources, producing 22,995 clips and 2.9 million frames across six tool--action pairs, with per-frame spatial annotations in nearly all clips (\cref{fig:overview,tab:comparison}).
These aligned annotations support recognition, temporal-boundary, tool-keypoint, and contact-region evaluation on the same actions.
The clips and annotations are available on \href{https://huggingface.co/datasets/hding2455/surgical-action-affordance}{Hugging Face}. The benchmark code is on \href{https://github.com/Ruifreddie/TACTUS}{GitHub}, and the \href{https://huggingface.co/abccbd/TACTUS-TactusNet}{model checkpoints} are publicly available.

To establish reference performance on these tasks, we introduce \net as the companion network and compare it with 13 published baselines on 11 Gold test videos.
With per-head encoder fine-tuning, \net achieves 95.8\% tool--action pair accuracy.
Across the evaluated methods, the best localization results reach 46.6\% recall at temporal IoU 0.7, 18.68 pixels of tool-point error at a resolution of $384\times384$, and 0.351 contact Dice.
Moreover, adding Silver to Gold training raises the default \net's pair accuracy from 93.8\% to 95.5\%.
By making complete tool--tissue interactions explicit targets for learning and evaluation, \tactus establishes a foundation for future surgical perception research supporting planning and control.

\section{Related work}
\label{sec:related}

\paragraph{Surgical datasets.}
Public surgical datasets support recognition and localization (\cref{tab:comparison}).
Cholec80, HeiChole, CholecT50, and ProstaTD provide workflow, instrument, and action annotations~\cite{twinanda2017endonet,wagner2023comparative,nwoye2022rendezvous,chen2026prostatd}, while CholecTrack20, PhaKIR, and SurgPose support instrument tracking and keypoint localization~\cite{nwoye2025cholectrack20,rueckert2026video,wu2025surgpose}.
However, these task-specific annotations only partially specify an interaction's temporal extent, instrument motion, and tissue involvement.
Larger collections, including OphNet and LEMON, broaden video and procedural coverage, but complete action-level spatio-temporal supervision remains limited~\cite{hu2024ophnet,che2026lemon}.
SurgAM extends spatial supervision to actionable tissue regions by expanding a single annotated pre-contact point into an affordance mask~\cite{song2026surgam}.
However, SurgAM provides this region annotation only for the pre-contact frame, leaving the tissue region's movement and deformation during the subsequent action unannotated.
To address these limitations, our new \tactus protocol uses each action or attempt as its annotation unit, jointly defining semantics, nested temporal windows, instrument trajectories, and tissue-region tracks (\cref{fig:formats}).

\paragraph{Model-assisted annotation.}
Model-assisted annotation provides established methods for extending sparse human input into dense supervision.
The SAM family generates masks from prompts~\cite{kirillov2023segment,ravi2025sam,carion2026sam}, while CoTracker propagates points across video frames~\cite{karaev2024cotracker,karaev2025cotracker3}.
SurgPose integrates SAM2-based propagation with frame-by-frame review, demonstrating how automated annotation can be combined with human quality control~\cite{wu2025surgpose}.
VISOR and InterHand2.6M further distinguish human annotations from model-generated labels, making their sources of supervision explicit~\cite{darkhalil2022epickitchens,moon2020interhand}.
Following these practices, \tactus uses point tracking to extend sparse spatial prompts and retains human semantic and temporal annotations in both tiers.
Gold combines manually placed spatial prompts with reviewed tracks, while Silver expands the collection through model-generated spatial prompts.
Moreover, the differing training benefits reported by COCONut and CoTracker3 motivate evaluating generated labels through downstream performance~\cite{deng2024coconut,karaev2025cotracker3}.
Accordingly, we assess the contribution of additional Silver annotations across recognition and localization tasks.

\begin{table}[t]
  \centering
  \captionsetup{skip=4pt}
  \caption{\textbf{Comparison with surgical video datasets.}
  \yes: the dataset's own labels answer the question (When: action start and end; How: tool keypoints or tool-keypoint trajectories; Where: tool--tissue contact location).
  Video and Hour refer to whole source videos, as reported or computed from durations or frame counts. --: not reported.
  Annot.: M, humans place or check labels, with optional model proposals or propagation. S, model-generated labels without individual review.
  \tactus Gold is M, and Silver is S.}
  \label{tab:comparison}
  \scriptsize
  \setlength{\tabcolsep}{2.2pt}
  \begin{tabular}{@{}lrrcccccc@{}}
    \toprule
    Dataset & Video & Hour & Annot. & Which & What & When & How & Where \\
    \midrule
    SurgAM~\cite{song2026surgam}            & --  & --   & S   & \no  & \yes & \no  & \no  & \yes \\
    PhaKIR~\cite{rueckert2026video}            & 8   & 5.4  & M   & \yes & \no  & \no  & \yes & \no  \\
    PSI-AVA~\cite{valderrama2022towards}      & 8   & 20   & M   & \yes & \yes & \no  & \no  & \no  \\
    GraSP~\cite{ayobi2025pixel}            & 13  & 32   & M   & \yes & \yes & \no  & \no  & \no  \\
    CholecTrack20~\cite{nwoye2025cholectrack20}     & 20  & 14   & M   & \yes & \no  & \no  & \no  & \no  \\
    ProstaTD~\cite{chen2026prostatd}   & 21  & --   & M   & \yes & \yes & \yes & \no  & \no  \\
    AutoLaparo~\cite{wang2022autolaparo}       & 21  & 23   & M   & \yes & \no  & \no  & \no  & \no  \\
    SurgPose~\cite{wu2025surgpose}         & 30  & --   & M   & \yes & \no  & \no  & \yes & \no  \\
    HeiChole~\cite{wagner2023comparative}          & 33  & 22   & M   & \yes & \yes & \yes & \no  & \no  \\
    M2CAI16~\cite{stauder2016tum,twinanda2016single}       & 41  & 26   & M   & \no  & \no  & \no  & \no  & \no  \\
    CholecT50~\cite{nwoye2022rendezvous}         & 50  & 28   & M   & \yes & \yes & \yes & \no  & \no  \\
    Cholec80~\cite{twinanda2017endonet}          & 80  & 51   & M   & \yes & \no  & \no  & \no  & \no  \\
    \midrule
    \textbf{\tactus (ours)}     & \textbf{141} & \textbf{66} & M+S & \yes & \yes & \yes & \yes & \yes \\
    \bottomrule
  \end{tabular}
\end{table}

\begin{figure}[t]
  \centering
  \includegraphics[width=0.90\linewidth]{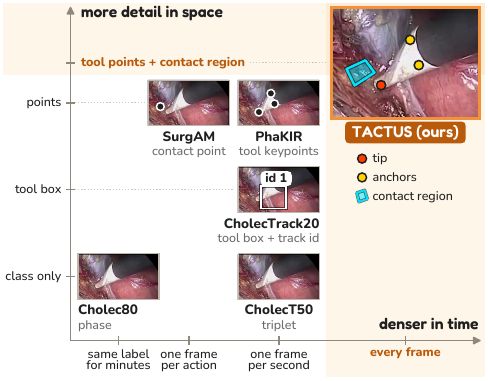}
  \captionsetup{skip=4pt}
  \caption{\textbf{Temporal and spatial scope of surgical annotations.}
  Each panel illustrates one annotation format on the same Cholec80 hook-dissection frame.
  CholecT50, CholecTrack20, and PhaKIR provide the labels shown at 1 fps.
  SurgAM uses a point on a single pre-contact frame to generate a region mask, whereas \tactus tracks tool keypoints and the contact region in every frame of the action window.}
  \label{fig:formats}
\end{figure}

\section{The \tactus dataset}
\label{sec:dataset}
We construct \tactus by applying a newly specified action annotation protocol to videos from five public sources.
The release contains 22,995 clips from 141 videos and 2.9 million frames, counting overlapping action windows once per video.
Its construction combines this protocol with two tiers of spatial supervision.

\subsection{Dataset construction}
\label{sec:construction}
\paragraph{Source selection and clip extraction.}
To identify suitable sources, we systematically screened 36 public surgical datasets and two YouTube collections for continuous video, clear instrument and tissue views, and relevant procedures and actions (Supp.~\ref{supp:B1}).
The five selected sources are Cholec80~\cite{twinanda2017endonet}, HeiChole~\cite{wagner2023comparative}, SurgVU~\cite{zia2025surgical}, CRCD~\cite{oh2024comprehensive,oh2025expanded}, and a collection of YouTube videos (\cref{tab:sources}).
Together, these sources cover clinical laparoscopic cholecystectomy, robot-assisted surgical training, and ex vivo cholecystectomy simulation.
First, annotators reviewed each complete video and identified actions with operative intent, including incomplete or failed attempts.
Each action attempt was then extracted as a separate clip and assigned tool and action labels, together with the two temporal spans defined below.
A new clip begins only when the action restarts, while pauses and temporary loss of visibility remain within the same attempt.

\paragraph{Gold quality control.}
For the 15,668 Gold clips, annotators placed spatial prompts on selected frames, and CoTracker3~\cite{karaev2025cotracker3} propagated the points throughout the window.
The annotation guide specifies two roles and two stages: annotators label the actions; reviewers inspect the complete videos, identify missed actions, and provide decisions and correction reasons; annotators then revise the returned clips; and reviewers finalize the remaining annotations.
Finally, the project added a core-team re-check of every finalized Gold clip.
Supp.~\ref{supp:B4} reports the annotation-stage review log, and Supp.~\ref{supp:E} gives the annotation rules.

\paragraph{Silver annotation.}
To reduce manual spatial prompting, Silver retains human annotation and review of tool classes, action classes, and temporal spans, while an early version of the \net Spatial Head generates the spatial prompts (Supp.~\ref{supp:B5}, \ref{supp:B6}).
The same tracker then propagates these prompts through each clip.
Silver contains 7,327 clips from SurgVU, and its spatial predictions are not reviewed point by point.

\paragraph{Data use and release.}
All source videos are publicly available, and no new patient or animal data were collected.
The release includes annotations, prompt points, review status, and clip boundaries. The source licenses govern distribution of the video clips (Supp.~\ref{supp:F}).

\subsection{Annotation definition}
\label{sec:definition}
Each clip is described through five linked dimensions, with all labels referring to the same action.
\paragraph{Which and What.}
These identify four tools (hook, grasper, scissors, and clipper) and four actions (dissect, grasp, clip, and cut), covering six annotated tool--action combinations.
\paragraph{When.}
Two nested temporal spans describe the action and its surrounding motion.
The \emph{action window} includes the tool's approach and completion or withdrawal, with boundaries determined by the tool-specific rules.
Within it, the \emph{interaction window} is defined by the visible events of the annotated action: for example, grasper grasp begins when the jaws start closing and ends when motion stops after retraction, while hook dissection begins with pulling or cauterizing and ends when the hook leaves the contact area.
Supp.~\ref{supp:A2}, \ref{supp:E1} give the corresponding events, and Supp.~\ref{supp:B3} gives the tolerances.
\paragraph{How.}
Three instrument-specific keypoints form per-frame trajectories throughout the action, with a visibility flag at each frame.
A hook has one tip point and two points at the head--shaft junction; jawed instruments have two tip points and one point at the jaw pivot or base.
Placement follows the tool-specific annotation rules in Supp.~\ref{supp:E3}.
\paragraph{Where.}
Four tissue-attached point tracks represent the contact region throughout the window, enclosing the affected tissue with a small buffer (Supp.~\ref{supp:B2}, \ref{supp:D2}).
During the interaction window, the region encloses contacted tissue; before and after it, the tracks follow the corresponding tissue through the window.
For both tool and tissue points, visibility flags determine their contribution to spatial supervision and evaluation.

\begin{table}[t]
  \centering
  \caption{\textbf{Released data sources.} Gold covers all five sources, and Silver comes from SurgVU only. Resolution and frame rate are those of the source videos used in TACTUS.}
  \label{tab:sources}
  \scriptsize
  \setlength{\tabcolsep}{2.2pt}
  \begin{tabular}{@{}l >{\raggedright\arraybackslash}p{2.3cm} l rr rr@{}}
    \toprule
    & & & \multicolumn{2}{c}{Gold} & \multicolumn{2}{c}{Silver} \\
    \cmidrule(lr){4-5}\cmidrule(l){6-7}
    Source & Setting & Video & Vid. & Clips & Vid. & Clips \\
    \midrule
    Cholec80~\cite{twinanda2017endonet}   & Clinical laparoscopic cholecystectomy & \makecell[l]{854$\times$480\\25 fps} & 30 & 6,058 & -- & -- \\
    SurgVU~\cite{zia2025surgical}    & Robot-assisted training, porcine model & \makecell[l]{1280$\times$720\\60 fps} & 32 & 4,281 & 40 & 7,327 \\
    YouTube              & Public laparoscopic cholecystectomy & \makecell[l]{$\le$1280$\times$720\\25--30 fps} & 21 & 2,766 & -- & -- \\
    HeiChole~\cite{wagner2023comparative}   & Clinical laparoscopic cholecystectomy & \makecell[l]{960$\times$540\\25 fps} & 10 & 1,799 & -- & -- \\
    CRCD~\cite{oh2024comprehensive,oh2025expanded}  & Robot-assisted pseudo-cholecystectomy, ex vivo porcine liver & \makecell[l]{1280$\times$792\\60 fps} & 8 & 764 & -- & -- \\
    \midrule
    Total & & & 101 & 15,668 & 40 & 7,327 \\
    \bottomrule
  \end{tabular}
\end{table}

\section{\net}
\label{sec:model}
\begin{figure}[t]
  \centering
  \includegraphics[width=\linewidth]{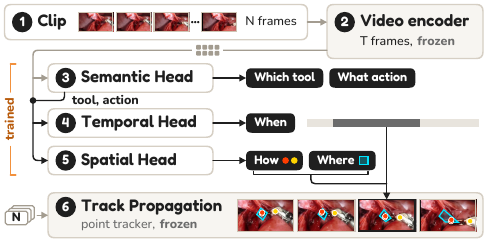}
  \caption{\textbf{The default \net predicts the five answers from shared tokens of a frozen video encoder.}
  The heads answer Which and What (Semantic), When (Temporal), and How and Where (Spatial) on $T$ sampled frames.
  For dense annotation, Track Propagation selects a key frame (black box) separately for tool and contact points and propagates them to all $N$ frames.}
  \label{fig:model}
\end{figure}
We introduce \net (\cref{fig:model}), a unified reference model for predicting the five annotated attributes of a surgical action.
Given a clip, we uniformly sample $T=16$ frames from its action window and encode them using a frozen V-JEPA~2.1 ViT-L~\cite{murlabadia2026vjepa} at $384\times384$ resolution.
The video tokens are shared by a Semantic Head for tool and action recognition, a Temporal Head for interval localization, and a Spatial Head for tool and contact points.
Tool and action classes condition both localization heads.

\paragraph{Semantic prediction.}
The annotated combinations are constrained by the dataset vocabulary, so the Semantic Head predicts a pair from the valid set $\mathcal K$.
A tool query and an action query cross-attend to the video tokens~\cite{carion2020detr}, with the tool prediction supplied to the action query as a prior.
Accordingly, the predicted pair is
\begin{equation}
\hat k=\operatorname*{arg\,max}_{k\in\mathcal K}
\left(\ell^{\mathrm t}_{\mathrm{tool}(k)}+\ell^{\mathrm a}_{\mathrm{act}(k)}\right),
\label{eq:pair}
\end{equation}
where $\ell^{\mathrm t}$ and $\ell^{\mathrm a}$ are the tool and action logits.
Pairs with a common tool or action share evidence, and the prediction belongs to $\mathcal K$.

\paragraph{Temporal localization.}
The Temporal Head predicts the interaction window within the action window.
Spatial pooling converts the video tokens into temporal features, and FiLM~\cite{perez2018film} incorporates the tool and action conditions.
Following anchor-free temporal localization~\cite{zhang2022actionformer,shi2023tridet}, the head predicts an actionness logit $A_t$, distances $\hat l_t$ and $\hat r_t$ to the interval boundaries, and a quality logit $Q_t$ at each sampled frame.
We select $t^*=\arg\max_t\sigma(A_t)\sigma(Q_t)$ and decode
\begin{equation}
\hat s=\frac{t^*-\hat l_{t^*}}{T-1},\qquad
\hat e=\frac{t^*+\hat r_{t^*}}{T-1},
\label{eq:interval}
\end{equation}
where $t\in\{0,\ldots,T-1\}$ and the distances are measured in sampling steps.
The decoded boundaries are normalized positions within the window.

\paragraph{Spatial localization.}
Instrument landmarks occupy a small image region, motivating coarse-to-fine prediction for fine-grained tool localization.
It first predicts tool points over the full frame and defines a crop around them.
The cropped frames are then encoded again, allowing separate decoders to estimate tool points and contact-region corners.
Finally, a refiner adjusts the tool points using image crops at the original resolution.
Each point decoder receives the tool and action conditions and predicts a confidence for every point.

For contact supervision, we derive a target from annotated tool-tip positions.
For sampled frame $t$ with source-frame index $f_t$, let $\mathcal F_t$ contain the frames within the annotated interaction window at or after $f_t$ that have a visible tip.
Specifically, the target map is
\begin{equation}
M_t(\mathbf g)=\max_{i\in\mathcal F_t}
\exp\!\left(-\frac{\|\mathbf g-\mathbf u_i\|^2}{2\sigma_g^2}\right),
\label{eq:hindsight}
\end{equation}
where $\mathbf u_i$ is the mean position of the visible tips in frame $i$ and $\sigma_g$ controls the Gaussian width.
An empty $\mathcal F_t$ is replaced by all frames with visible tips in the interval; clips without any such frame receive no contact target.
From these tip positions, we derive four corner targets (Supp.~\ref{supp:C1}).
This target describes tool-tip activity, whereas the benchmark evaluates overlap with the human-defined tissue regions (Supp.~\ref{supp:C3}).

\paragraph{Training and inference.}
In the default configuration, the video encoder remains frozen, and the heads are trained separately using its shared features.
Ground-truth tool and action labels condition the Temporal and Spatial Heads.
At inference, these conditions come from the Semantic Head or supplied labels.
For dense annotation, we select the most confident sampled frame within the interaction window separately for tool and contact points, and propagate the prompts using CoTracker3 (Supp.~\ref{supp:C1}).
Model-assisted annotation uses the human-annotated interaction window for this selection.
For benchmark evaluation, spatial scores use the direct predictions on sampled frames, before propagation.
Supp.~\ref{supp:C} provides the architecture, losses, and optimization details.

\section{Experiments}
\label{sec:experiments}
\begin{table*}[t]
  \centering
  \caption{\textbf{No method leads every question on the Gold test set.}
  The learned methods are trained on the Gold training set, and the floors have no trainable parameters (the pair-mean interval uses statistics of that set).
  When, How, and Where receive the true tool and action.
  In (a), each baseline predicts one of the six pairs, and its Which and What follow from this pair.
  In (c), the hull of the four predicted contact points is the Where region of the keypoint methods.
  Pretraining: the public weights that the backbone starts from.
  $\dagger$: frozen; all other backbones are fine-tuned (\net: shared by all heads or one per head group; \cref{sec:ablation}).
  $^*$: its pretraining included two of our eleven test videos.
  /:~no backbone or pretrained weights (floor). --:~not applicable.
  mAcc: mean accuracy over the six pairs.
  R@0.7: share of intervals with tIoU $\geq 0.7$.
  $>$32\,px: share of tool points off by more than 32 pixels at $384\times384$.
  Supp.~\ref{supp:D5}, \ref{supp:D7} gives the baseline and fine-tuning configurations.}
  \label{tab:results}
  \footnotesize
  \setlength{\tabcolsep}{5pt}
  \begin{tabular}{@{}lll*{4}{>{\centering\arraybackslash}p{1.8cm}}@{}}
    \toprule
    \multicolumn{3}{@{}l}{\textit{(a) Classification}} & Which & What & \multicolumn{2}{c}{Pair} \\
    \cmidrule(lr){4-4}\cmidrule(lr){5-5}\cmidrule(l){6-7}
    & Backbone & Pretraining & Acc (\%)$\uparrow$ & Acc (\%)$\uparrow$ & Acc (\%)$\uparrow$ & mAcc (\%)$\uparrow$ \\
    \midrule
    Majority class                   & /                          & /              & 68.4 & 77.2 & 68.4 & 16.7 \\
    TSN~\cite{wang2016tsn}           & ResNet-50                  & ImageNet-1k    & 84.7 & 81.4 & 78.4 & 56.3 \\
    TSM~\cite{lin2019tsm}            & ResNet-50                  & SSv2           & 87.8 & 90.3 & 85.1 & 56.5 \\
    VideoMAE~\cite{tong2022videomae} & ViT-B/16                   & VideoMAE, SSv2 & 94.2 & 96.2 & 92.8 & 73.3 \\
    X-CLIP~\cite{ni2022xclip}        & ViT-B/16                   & CLIP, K400     & 91.3 & 90.6 & 87.1 & 68.7 \\
    Endo-FM\smash{$^*$}~\cite{wang2023endofm} & ViT-B/16                   & Endo-FM        & 85.1 & 82.9 & 80.0 & 42.5 \\
    \net (ours)                      & ViT-L/16\smash{$^\dagger$} & V-JEPA~2.1     & 96.3 & 96.5 & 93.8 & 82.7 \\
    \quad fine-tuned, shared         & ViT-L/16                   & V-JEPA~2.1     & 96.3 & 96.9 & 94.1 & 84.9 \\
    \quad fine-tuned, per head       & 3$\times$ViT-L/16          & V-JEPA~2.1     & 97.6 & 97.8 & 95.8 & 89.2 \\
    \bottomrule
    \addlinespace[0.6em]
    \toprule
    \multicolumn{3}{@{}l}{\textit{(b) Temporal localization}} & \multicolumn{4}{c}{When} \\
    \cmidrule(l){4-7}
    & Backbone & Pretraining & \multicolumn{2}{c}{tIoU$\uparrow$} & \multicolumn{2}{c}{R@0.7 (\%)$\uparrow$} \\
    \midrule
    Whole window                              & /                          & /          & \multicolumn{2}{c}{0.454} & \multicolumn{2}{c}{15.4} \\
    Pair-mean interval                        & /                          & /          & \multicolumn{2}{c}{0.569} & \multicolumn{2}{c}{36.1} \\
    ActionFormer~\cite{zhang2022actionformer} & ViT-L/16\smash{$^\dagger$} & V-JEPA~2.1 & \multicolumn{2}{c}{0.624} & \multicolumn{2}{c}{46.6} \\
    TriDet~\cite{shi2023tridet}               & ViT-L/16\smash{$^\dagger$} & V-JEPA~2.1 & \multicolumn{2}{c}{0.616} & \multicolumn{2}{c}{45.5} \\
    QD-DETR~\cite{moon2023qddetr}             & ViT-L/16\smash{$^\dagger$} & V-JEPA~2.1 & \multicolumn{2}{c}{0.599} & \multicolumn{2}{c}{44.3} \\
    MS-TCN~\cite{farha2019mstcn}              & ViT-L/16\smash{$^\dagger$} & V-JEPA~2.1 & \multicolumn{2}{c}{0.607} & \multicolumn{2}{c}{43.3} \\
    \net (ours)                               & ViT-L/16\smash{$^\dagger$} & V-JEPA~2.1 & \multicolumn{2}{c}{0.592} & \multicolumn{2}{c}{40.7} \\
    \quad fine-tuned, shared                  & ViT-L/16                   & V-JEPA~2.1 & \multicolumn{2}{c}{0.592} & \multicolumn{2}{c}{40.8} \\
    \quad fine-tuned, per head                & 3$\times$ViT-L/16          & V-JEPA~2.1 & \multicolumn{2}{c}{0.592} & \multicolumn{2}{c}{40.4} \\
    \bottomrule
    \addlinespace[0.6em]
    \toprule
    \multicolumn{3}{@{}l}{\textit{(c) Keypoints and regions}} & \multicolumn{2}{c}{How} & \multicolumn{2}{c}{Where} \\
    \cmidrule(lr){4-5}\cmidrule(l){6-7}
    & Backbone & Pretraining & ADE (px)$\downarrow$ & $>$32\,px (\%)$\downarrow$ & \multicolumn{2}{c}{Dice$\uparrow$} \\
    \midrule
    DARK~\cite{zhang2020dark}         & HRNet-W32                  & ImageNet-1k      & 23.88 & 24.4 & \multicolumn{2}{c}{0.347} \\
    ViTPose~\cite{xu2022vitpose}      & ViT-B/16                   & MAE, ImageNet-1k & 23.64 & 24.0 & \multicolumn{2}{c}{0.339} \\
    DSTA~\cite{he2024dsta}            & ResNet-50                  & ImageNet-1k      & 25.70 & 25.2 & \multicolumn{2}{c}{0.312} \\
    SegFormer~\cite{xie2021segformer} & MiT-B2                     & ImageNet-1k      & --          & --         & \multicolumn{2}{c}{0.351} \\
    \net (ours)                       & ViT-L/16\smash{$^\dagger$} & V-JEPA~2.1       & 19.15       & 20.3       & \multicolumn{2}{c}{0.242} \\
    \quad fine-tuned, shared          & ViT-L/16                   & V-JEPA~2.1       & 19.10       & 20.3       & \multicolumn{2}{c}{0.244} \\
    \quad fine-tuned, per head        & 3$\times$ViT-L/16          & V-JEPA~2.1       & 18.68       & 19.9       & \multicolumn{2}{c}{0.244} \\
    \bottomrule
  \end{tabular}
\end{table*}
\subsection{Tasks, data splits and evaluation}
\label{sec:metrics}\label{sec:stats}
We systematically evaluate \net and 13 published methods across the five annotation dimensions.
The benchmark uses 15,499 Gold clips, split by video into 12,372 training, 1,604 validation, and 1,523 test clips (Supp.~\ref{supp:D1}).
All checkpoints are selected on the validation set.
All learned models in the main comparison train on Gold; the default \net uses a frozen video encoder.
Localization uses ground-truth tool and action labels; interaction windows, tool keypoints, and contact regions are scored separately.
We further evaluate training with additional Silver annotations. We also assess encoder fine-tuning.

\paragraph{Tool and action recognition.}
We report top-1 accuracy for the tool, action, and pair, together with mean accuracy across the six pairs.
The baselines are TSN~\cite{wang2016tsn}, TSM~\cite{lin2019tsm}, VideoMAE~\cite{tong2022videomae}, X-CLIP~\cite{ni2022xclip}, and Endo-FM~\cite{wang2023endofm}, with a majority-class reference.
\paragraph{Temporal localization.}
We report temporal IoU (tIoU) and recall at a tIoU threshold of 0.7.
ActionFormer~\cite{zhang2022actionformer}, TriDet~\cite{shi2023tridet}, QD-DETR~\cite{moon2023qddetr}, and MS-TCN~\cite{farha2019mstcn} use features from the same frozen encoder as \net.
Whole-window and pair-mean intervals provide references requiring no model training.
\paragraph{Tool-keypoint localization.}
We report average displacement error (ADE) and the share of visible tool points with errors above 32 pixels.
Spatial metrics use 16 uniformly sampled frames in a $384\times384$ coordinate system.
DARK~\cite{zhang2020dark}, ViTPose~\cite{xu2022vitpose}, and DSTA~\cite{he2024dsta} serve as keypoint baselines.
\paragraph{Contact-region estimation.}
We measure overlap with the annotated tissue regions using Dice.
For keypoint methods, the region is the convex hull of the contact points; SegFormer~\cite{xie2021segformer} predicts a region mask.
Metric definitions and baseline adaptations are given in Supp.~\ref{supp:D2}, \ref{supp:D5}.
\label{sec:baselines}
\paragraph{Baseline implementation.}
To apply this protocol, we initialize the baselines with public pretrained backbone weights and follow their published training recipes with the adaptations listed in Supp.~\ref{supp:D5}.
Specifically, classification baselines predict the six pairs, from which tool and action labels are read.
For spatial localization, the keypoint baselines predict three tool points per tool and four contact points per pair; training and evaluation select the outputs associated with the annotated classes.
These models operate on the complete content frame, using a fixed crop per video to remove borders and interface panels.
The temporal baselines use the true pair's output channel, while QD-DETR receives a fixed text query for that pair.
The interaction window supplies the temporal supervision target.

\begin{figure*}[!t]
  \centering
  \includegraphics[width=0.97\textwidth]{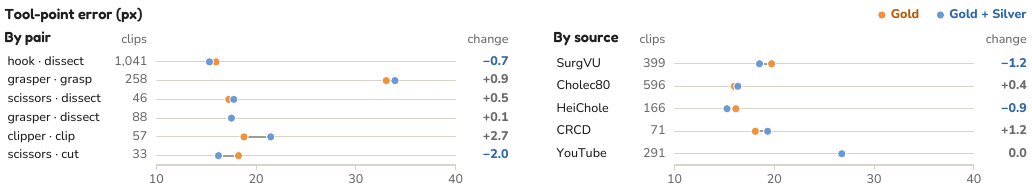}
  \captionsetup{skip=4pt}
  \caption{\textbf{Tool-point localization by pair and source.} Left: by pair. Right: by source. Orange: Gold only; blue: Gold + Silver. Change: difference in pixels (negative: Silver lowers the error). Results use the 1,523 Gold test clips.}
  \label{fig:pairsource}
\end{figure*}

\subsection{Main results}
\label{sec:results}
\Cref{tab:results} reports results across the five tasks.
Qualitative examples and subgroup analyses appear in the supplement.
\paragraph{Tool and action recognition.}
The default \net achieves 93.8\% pair accuracy, compared with 92.8\% for VideoMAE, the strongest external classifier.
Its mean pair accuracy is 82.7\%, compared with 73.3\% for VideoMAE.
\paragraph{Temporal localization.}
ActionFormer achieves the highest tIoU of 0.624 and a recall of 46.6\% at the 0.7 threshold.
All four temporal baselines outperform the default \net Temporal Head, which reaches 0.592 tIoU.
However, the pair-mean interval already achieves 0.569 without observing the video, leaving a margin of 0.055 for ActionFormer.
\paragraph{Tool-keypoint localization.}
The default \net achieves 19.15 pixels ADE, compared with 23.64 for ViTPose.
Despite this lower average error, 20.3\% of its visible tool points remain more than 32 pixels from the annotated positions.
\paragraph{Contact-region estimation.}
SegFormer achieves the highest Dice of 0.351, followed by the keypoint baselines at 0.312--0.347 and the default \net at 0.242.
These methods also use different supervision targets.
The baselines learn human-defined tissue regions, whereas \net learns tip-derived targets, making the observed differences a combined effect of architecture and supervision (Supp.~\ref{supp:C3}).
Within \net, Dice increases from 0.096 on frames with two visible contact points to 0.306 with four (Supp.~\ref{supp:D3}).
This visibility-dependent gap is consistent with the geometry of the annotated regions, as two visible points produce thin targets whose overlap is particularly sensitive to spatial offsets.

\begin{table*}[t]
  \centering
  \caption{\textbf{Ablations and analyses of \net on the Gold test set.}
  Backbone (a), spatial stages (b), training data and encoder (c), and class conditions (d).
  Shaded rows: the setting of \net in \cref{tab:results}.
  $\dagger$, $^*$, and metrics as in \cref{tab:results}.}
  \label{tab:ablation}
  \footnotesize
  \setlength{\tabcolsep}{3.5pt}
  \renewcommand{\arraystretch}{1.00}
  \newcommand{\refrow}{\rowcolor{black!8}}
  \sbox{\ablA}{\begin{tabular}[t]{lll*{5}{c}}
      \multicolumn{8}{@{}l}{\textbf{(a) Backbone}} \\
      \toprule
      & & & Which & What & Pair & When & How (stage 1) \\
      \cmidrule(lr){4-4}\cmidrule(lr){5-5}\cmidrule(lr){6-6}\cmidrule(lr){7-7}\cmidrule(lr){8-8}
      Backbone & Pretraining & Input & Acc$\uparrow$ & Acc$\uparrow$ & Acc$\uparrow$ & tIoU$\uparrow$ & ADE$\downarrow$ \\
      \midrule
      ViT-L/16$^\dagger$         & VideoMAE, K400~\cite{tong2022videomae}   & $16{\times}224^2$ & 92.0 & 94.1 & 89.7 & 0.588 & 21.76 \\
      ViT-B/16$^\dagger$         & Endo-FM$^*$~\cite{wang2023endofm}          & $16{\times}224^2$ & 83.9 & 86.5 & 79.7 & 0.586 & 30.60 \\
      ViT-L/16$^\dagger$         & V-JEPA~2~\cite{assran2025vjepa2}         & $16{\times}256^2$ & 93.1 & 95.1 & 90.6 & 0.589 & 21.48 \\
      \refrow ViT-L/16$^\dagger$ & V-JEPA~2.1~\cite{murlabadia2026vjepa}    & $16{\times}384^2$ & 96.3 & 96.5 & 93.8 & 0.592 & 20.24 \\
      \bottomrule
    \end{tabular}}%
  \sbox{\ablB}{\begin{tabular}[t]{lccc}
      \multicolumn{4}{@{}l}{\textbf{(b) Spatial Head}} \\
      \toprule
      & \multicolumn{3}{c}{How} \\
      \cmidrule(lr){2-4}
      Stage & ADE$\downarrow$ & Median$\downarrow$ & $>$32\,px$\downarrow$ \\
      \midrule
      Stage 1, full frame & 20.24 & 12.5 & 20.0 \\
      + zoom (stage 2)    & 19.42 & 10.4 & 20.0 \\
      \refrow + refiner   & 19.15 & 9.9  & 20.3 \\
      \bottomrule
    \end{tabular}}%
  \sbox{\ablC}{\begin{tabular}[t]{ll*{6}{c}}
      \multicolumn{8}{@{}l}{\textbf{(c) Training data and encoder}} \\
      \toprule
      & & Which & What & Pair & When & How & Where \\
      \cmidrule(lr){3-3}\cmidrule(lr){4-4}\cmidrule(lr){5-5}\cmidrule(lr){6-6}\cmidrule(lr){7-7}\cmidrule(lr){8-8}
      Data & Encoder & Acc$\uparrow$ & Acc$\uparrow$ & Acc$\uparrow$ & tIoU$\uparrow$ & ADE$\downarrow$ & Dice$\uparrow$ \\
      \midrule
      \refrow Gold  & frozen & 96.2 & 96.5 & 93.8 & 0.592 & 19.15 & 0.242 \\
      Gold + Silver & frozen & 98.2 & 96.1 & 95.5 & 0.587 & 18.93 & 0.244 \\
      Silver only & frozen & 47.7 & 91.1 & 43.5 & 0.558 & 27.10 & 0.271 \\
      \addlinespace[2pt]
      Gold & frozen, +10 epochs   & 96.1 & 96.8 & 94.0 & 0.591 & 19.08 & 0.242 \\
      Gold & fine-tuned, shared   & 96.1 & 96.9 & 94.1 & 0.592 & 19.10 & 0.244 \\
      Gold & fine-tuned, per head & 97.6 & 97.8 & 95.8 & 0.592 & 18.68 & 0.244 \\
      \bottomrule
    \end{tabular}}%
  \sbox{\ablD}{\begin{tabular}[t]{lccc}
      \multicolumn{4}{@{}l}{\textbf{(d) Conditions}} \\
      \toprule
      & When & How & Where \\
      \cmidrule(lr){2-2}\cmidrule(lr){3-3}\cmidrule(lr){4-4}
      Tool and action & tIoU$\uparrow$ & ADE$\downarrow$ & Dice$\uparrow$ \\
      \midrule
      \refrow True (protocol) & 0.592 & 19.15 & 0.242 \\
      Predicted pair          & 0.590 & 19.85 & 0.241 \\
      \bottomrule
    \end{tabular}}%
  \setlength{\ablL}{\wd\ablA}\ifdim\wd\ablC>\ablL\setlength{\ablL}{\wd\ablC}\fi
  \setlength{\ablR}{\wd\ablB}\ifdim\wd\ablD>\ablR\setlength{\ablR}{\wd\ablD}\fi
  \makebox[\ablL][l]{\subtab{\ablA}{Frozen checkpoints with retrained heads and 16 frames at their pretraining resolution.
    Encoder size and resolution vary. How uses stage 1; Where is omitted. Input: frames $\times$ image size.}}\hfill
  \makebox[\ablR][l]{\subtab{\ablB}{Successive outputs of one trained Spatial Head, without retraining.
    Median: median tool-point error in pixels.}}
  \par\vspace{6pt}
  \makebox[\ablL][l]{\subtab{\ablC}{The last three rows add ten training epochs.
    Which and What use head logits. Panel (a) and \cref{tab:results} use the predicted pair.}}\hfill
  \makebox[\ablR][l]{\subtab{\ablD}{The same trained heads receive true classes or the predicted pair (93.8\% correct). Both rows use all test clips.}}
\end{table*}
\subsection{Ablation studies}
\label{sec:ablation}
We examine additional training data, encoder configuration, spatial refinement, and class conditions (\cref{tab:ablation}).
\paragraph{The complementary roles of Gold and Silver.}
\label{sec:silver}
To assess the contribution of Silver annotations, we compare Gold-only, Silver-only, and joint training under the same evaluation protocol (\cref{tab:ablation}(c)).
On SurgVU, Silver-only training achieves 18.67 pixels ADE, compared with 19.73 pixels for Gold-only training, showing that Silver provides useful supervision for tool localization within its source domain.
Moreover, adding Silver clips and their human class labels to Gold training increases pair accuracy from 93.8\% to 95.5\%.
Together, the source-domain localization result and the recognition gains support the training value of Silver.
Gold provides multi-source supervision, while Silver expands the collection through model-assisted spatial annotation (Supp.~\ref{supp:D3}).
\label{sec:analysis}
\paragraph{Results by pair and source.}
\Cref{fig:pairsource} reveals substantial variation in tool-point localization across both maneuvers and video sources.
Grasper grasp has the largest error, at about 33 pixels, compared with 16--19 pixels for the other pairs.
Similarly, the error on YouTube clips reaches about 27 pixels, compared with 16--20 pixels for the other sources.
Against this variation, adding Silver changes each group's ADE by less than 3 pixels.
Its largest source-level improvement occurs on SurgVU, which supplies the Silver clips, suggesting that the additional annotations provide their clearest localization benefit within the represented source domain.

\paragraph{The choice of video encoder.}
We compare frozen VideoMAE, Endo-FM, V-JEPA~2~\cite{assran2025vjepa2}, and V-JEPA~2.1 configurations with retrained heads.
At the tested model sizes and resolutions, V-JEPA~2.1 gives the strongest recognition and tool-point results, supporting its use as the default encoder.
Temporal performance ranges from 0.586 to 0.592 tIoU.
\paragraph{Shared and per-head fine-tuning.}
We compare shared and per-head encoder fine-tuning with a frozen control, all trained for ten additional epochs.
Per-head fine-tuning raises pair accuracy from 94.0\% to 95.8\% and reduces ADE from 19.08 to 18.68 pixels.
The gain also appears in mean pair accuracy, which weights all six pairs equally: 83.0\% for the frozen control, 84.9\% for shared fine-tuning, and 89.2\% for separate encoders (Supp.~\ref{supp:D7}).
Compared with equally extended head training, per-head encoder adaptation improves recognition and tool-point localization using three encoder instances (Supp.~\ref{supp:D7}).
\paragraph{Spatial refinement.}
Across successive outputs of the same trained Spatial Head, zoom reduces ADE from 20.24 to 19.42 pixels, and the refiner further reduces it to 19.15.
The median error falls from 12.5 to 9.9 pixels, while the share above 32 pixels remains near 20\%.
These reductions support local refinement for improving instrument-landmark precision.
\paragraph{Predicted class conditions.}
To measure error propagation, we replace ground-truth classes with the predicted pair.
ADE rises from 19.15 to 19.85 pixels, while tIoU and Dice decrease by 0.002 and 0.001.
The increased tool-point error is consistent with instrument-specific landmark definitions, linking semantic predictions to subsequent spatial estimation.

\section{Conclusion}
\label{sec:conclusion}
We presented \tactus, a dataset and benchmark built on a unified annotation protocol for complete surgical actions.
The protocol connects action semantics, nested temporal windows, instrument trajectories, and tissue-region tracks, establishing explicit learning targets for understanding how instruments interact with tissue over time.
Through a scalable Gold--Silver construction pipeline, we created 22,995 action clips from 141 videos across five public sources.
We further introduced \net and a common evaluation framework for semantic, temporal, and spatial prediction.
Our experiments demonstrate the benefits of spatial refinement and the complementary training value of Gold and Silver annotations.
Together, these contributions provide a foundation for developing surgical perception models that connect action recognition with execution timing, instrument motion, and tissue involvement.
Future work will extend the protocol to additional procedures and instruments and expand Silver across more diverse video sources.
By broadening the scale and coverage of this supervision, we aim to advance surgical action understanding and pave the way for future perception-guided surgical automation.

\endgroup
\clearpage
\maketitlesupplementary
\pdfbookmark[0]{Appendix: Supplementary Material}{supplementary-material}
\appendix
\crefalias{section}{appendix}\crefalias{subsection}{subappendix}%
\crefname{appendix}{Supp.}{Supp.}\Crefname{appendix}{Supp.}{Supp.}
\crefname{subappendix}{Supp.}{Supp.}\Crefname{subappendix}{Supp.}{Supp.}
\raggedbottom
\renewcommand{\topfraction}{0.95}\renewcommand{\bottomfraction}{0.6}\renewcommand{\textfraction}{0.05}
\renewcommand{\floatpagefraction}{0.85}\renewcommand{\dbltopfraction}{0.95}\renewcommand{\dblfloatpagefraction}{0.85}
\setlength{\textfloatsep}{12pt plus 2pt minus 4pt}\setlength{\dbltextfloatsep}{12pt plus 2pt minus 4pt}\setlength{\floatsep}{10pt plus 2pt minus 2pt}\setlength{\dblfloatsep}{10pt plus 2pt minus 2pt}\setlength{\intextsep}{10pt plus 2pt minus 2pt}
\makeatletter\setlength\@fptop{0pt}\setlength\@fpsep{12pt plus 2pt}\setlength\@fpbot{0pt plus 1fil}\setlength\@dblfptop{0pt}\setlength\@dblfpsep{12pt plus 2pt}\setlength\@dblfpbot{0pt plus 1fil}\makeatother
\setcounter{topnumber}{4}\setcounter{dbltopnumber}{3}\setcounter{totalnumber}{6}
\setlength{\abovedisplayskip}{8pt plus 2pt minus 2pt}
\setlength{\belowdisplayskip}{8pt plus 2pt minus 2pt}
\setlength{\abovedisplayshortskip}{3pt plus 2pt}
\setlength{\belowdisplayshortskip}{5pt plus 2pt minus 2pt}
\setcounter{figure}{0}\renewcommand{\thefigure}{S\arabic{figure}}\renewcommand{\theHfigure}{supp.\arabic{figure}}
\setcounter{table}{0}\renewcommand{\thetable}{S\arabic{table}}\renewcommand{\theHtable}{supp.\arabic{table}}
\setcounter{equation}{0}\renewcommand{\theequation}{S\arabic{equation}}\renewcommand{\theHequation}{supp.\arabic{equation}}
\noindent This supplement details how \tactus is defined, constructed, and evaluated.
We first describe the source videos and action definitions (\cref{supp:A}), followed by the construction pipeline and annotation-quality evidence (\cref{supp:B}).
We then present the model implementation (\cref{supp:C}) and evaluation protocols and additional results (\cref{supp:D}).
The operational annotation rules (\cref{supp:E}) and release contents (\cref{supp:F}) provide the remaining specifications.
\section{Surgical background}
\label{supp:A}

The surgical context below explains the visual differences among the sources and the tissue interactions represented by the tool--action pairs. The timing analysis in \cref{supp:A3} characterizes the pair- and tier-specific distributions relevant to When.

\subsection{Surgical scenes of the five sources}
\label{supp:A1}

The sources selected in \cref{sec:construction} differ in tissue condition, instrument design, and image appearance.
Cholec80, HeiChole, and the YouTube collection show laparoscopic cholecystectomy performed through small incisions, while SurgVU shows robot-assisted surgical training on porcine models. CRCD provides endoscopic views of robot-assisted cholecystectomy simulated on an ex vivo porcine liver placed on a table.
The following source characteristics account for these visual differences.

\paragraph{Clinical laparoscopy.}
Cholec80, HeiChole, and YouTube depict cholecystectomy, the removal of the gallbladder, in patients.
These clinical scenes include bleeding from living tissue, smoke produced by electrical cautery, and rapid camera motion~\cite{twinanda2017endonet}.
Their acquisition conditions also vary: Cholec80 originates from one hospital~\cite{twinanda2017endonet}, whereas HeiChole spans three centers, two camera systems, and several instrument vendors~\cite{wagner2023comparative}.
Some YouTube frames additionally contain overlaid text, logos, or camera menus.

\paragraph{Robotic simulation on ex vivo tissue.}
CRCD records simulated cholecystectomy with a first-generation da Vinci robot on an ex vivo porcine liver and gallbladder, removed from a pig~\cite{oh2024comprehensive,oh2025expanded}.
The absence of blood flow distinguishes this tissue from clinical scenes, while the similar colors of the liver and gallbladder can limit their visual separation~\cite{oh2024comprehensive}.
\tactus uses the left image from CRCD's stereo endoscope.

\paragraph{Robotic training on porcine models.}
SurgVU provides the second robotic source, with trainees and experts performing standard surgical steps on porcine models~\cite{zia2025surgical}.
Up to three instruments can appear in a SurgVU frame. In its tool-detection validation set, embedded interface information is blurred~\cite{zia2025surgical}.

\paragraph{Instruments.}
Laparoscopic instruments have straight shafts, while many da Vinci instruments have an articulated distal wrist~\cite{zia2025surgical}.
This articulation changes the orientation of the working tip relative to the shaft and informs the tool-specific keypoint definitions (\cref{supp:E3}).

\subsection{Tool--tissue interactions and boundaries}
\label{supp:A2}

\begin{table*}[t]
  \centering
  \caption{\textbf{Tissue interactions and interaction-window events for the six tool--action pairs.}
  Four pairs act on tissue and two on ducts and vessels. Rows are ordered by clip count in the full release, from largest to smallest.}
  \label{tab:suppA-pairs}
  \footnotesize
  \setlength{\tabcolsep}{4pt}
  \renewcommand{\arraystretch}{1.15}
  \begin{tabular}{@{}l >{\raggedright\arraybackslash}p{1.9cm} >{\raggedright\arraybackslash}p{5.3cm} >{\raggedright\arraybackslash}p{3.5cm} >{\raggedright\arraybackslash}p{3.0cm}@{}}
    \toprule
    Pair & Acts on & Effect on tissue & Interaction starts when & Interaction ends when \\
    \midrule
    hook dissect     & Tissue, including fat & Retracts or cauterizes tissue to separate tissue layers and expose the structures beneath & The hook starts to retract or cauterize, whichever comes first & The hook leaves the contact area \\
    grasper grasp    & Tissue, organs & Holds tissue and retracts or stabilizes it, to improve the view and access or to support another action such as dissection & The jaws start to close & The tool stops moving after the retraction \\
    scissors dissect & Tissue & Separates tissue; the protocol gives this pair no events of its own & The tip begins to work on or deliberately touch tissue & The tip leaves the contact area or retracts \\
    grasper dissect  & Tissue, including fat & The closed tips enter the tissue shallowly, and the opening jaws then separate it bluntly & The closed jaws enter the tissue & The jaws stop opening \\
    clipper clip     & Ducts, vessels & Places a clip on a duct or vessel to stop bleeding or to close it off & Clipping starts: the jaws close & The clipper resets and lets go \\
    scissors cut     & Ducts, vessels & Closes the blades across a clipped duct or vessel & Cutting starts & Cutting ends \\
    \bottomrule
  \end{tabular}
\end{table*}

\Cref{tab:suppA-pairs} specifies the visible events that delimit each pair's interaction window, extending the general definition in \cref{sec:definition}.
Scissors dissect follows the general rule without additional pair-specific events.

\subsection{Action and interaction window lengths}
\label{supp:A3}

\suppressfloats[t]
\begin{figure}[t]
  \centering
  \includegraphics[width=\linewidth]{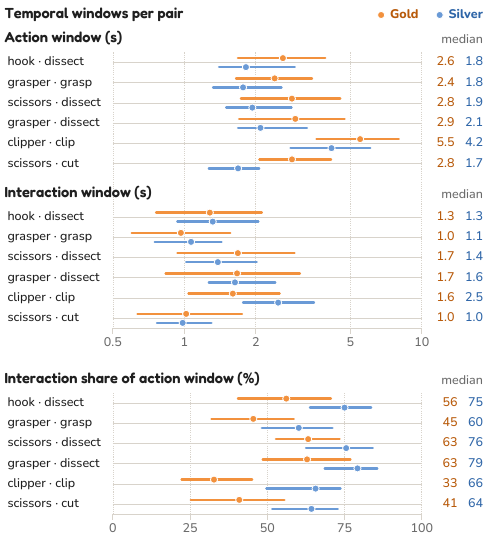}
  \caption{\textbf{Pair-specific timing distributions and shorter action windows in Silver.}
  Top and middle: action-window and interaction-window lengths in seconds, shown on a shared logarithmic axis.
  Bottom: the fraction of the action window occupied by the interaction window.
  Dots and numerical labels indicate medians; bars span the 25th to 75th percentiles.
  Orange: Gold. Blue: Silver.}
  \label{fig:suppA-lengths}
\end{figure}

The pair-specific distributions in \cref{fig:suppA-lengths} provide a temporal cue for When (\cref{sec:stats}).
Silver action windows are shorter for every pair, including when compared only with Gold clips from SurgVU, Silver's sole source. Both tiers follow the shared action and interaction definitions, with operational rules and tolerances specified in \cref{tab:supp-E-gold,tab:supp-E-silver,supp:B3}.
Median interaction-window lengths differ by at most 0.3\,s, except for clipper clip, whose Silver median is 0.9\,s longer.
Consequently, the interaction window occupies a larger fraction of the Silver action window for every pair. Adding Silver changes this distribution during Temporal Head training, while the measured tIoU remains nearly unchanged (\cref{sec:silver}).

\section{Dataset construction}
\label{supp:B}

The construction pipeline integrates source selection, spatial annotation, track propagation, and quality control (\cref{sec:construction,fig:pipeline}); \cref{supp:E} provides the operational rules.

\subsection{Source survey}
\label{supp:B1}

\begin{table*}[p]
  \centering
  \caption{\textbf{Source selection from the surveyed datasets and video collections.}
  Five sources satisfy all selection criteria. Shortlisted entries receive a survey verdict; pooled entries remained candidates for annotation.
  For sources excluded from the shortlist, the reason identifies the property that fails a criterion in \cref{sec:construction}.}
  \label{tab:B-survey}
  \footnotesize
  \setlength{\tabcolsep}{4pt}
  \begin{tabular}{@{}>{\raggedright}p{5.4cm} >{\raggedright}p{4.9cm} >{\raggedright}p{2.3cm} >{\raggedright\arraybackslash}p{3.8cm}@{}}
    \toprule
    Dataset & Procedure & Survey verdict & Reason \\
    \midrule
    \multicolumn{4}{@{}l}{\textit{Used: four datasets and one YouTube collection}} \\
    Cholec80~\cite{twinanda2017endonet}, HeiChole~\cite{wagner2023comparative}, CRCD~\cite{oh2024comprehensive,oh2025expanded}, SurgVU~\cite{zia2025surgical}, YouTube cholecystectomy collection
      & Cholecystectomy, its robot-assisted simulation, and robot-assisted training (\cref{tab:sources}) & Suitable & Meets all criteria \\
    \midrule
    \multicolumn{4}{@{}l}{\textit{Shortlisted, not used: nine datasets}} \\
    GraSP~\cite{ayobi2025pixel}                     & Robot-assisted prostatectomy                  & Suitable, pooled  & Other procedure \\
    SAR-RARP50~\cite{psychogyios2024sarrarp50}      & Robot-assisted prostatectomy                  & Suitable, pooled  & Mostly suturing, not one of our actions \\
    AutoLaparo~\cite{wang2022autolaparo}            & Laparoscopic hysterectomy                     & Undecided, pooled & Other procedure \\
    MultiBypass140~\cite{lavanchy2024challenges}    & Laparoscopic gastric bypass                   & Undecided         & Other procedure \\
    PhaKIR~\cite{rueckert2026video}                 & Laparoscopic cholecystectomy                  & Undecided         & Proposed after the Gold videos were labeled \\
    SARAS-MESAD~\cite{bawa2021saras}                & Robot-assisted prostatectomy, real and phantom & Rejected         & Too sparse in time \\
    SARAS-ESAD~\cite{bawa2021saras}                 & Robot-assisted prostatectomy                  & Rejected          & Too small \\
    SurgT~\cite{cartucho2024surgt}                  & Robot-assisted, porcine and human cases       & Rejected          & Too small \\
    SurgicalActions160~\cite{schoeffmann2018video}  & Gynecologic laparoscopy                       & Rejected          & Too small \\
    \midrule
    \multicolumn{4}{@{}l}{\textit{Not shortlisted: 23 datasets and one YouTube collection}} \\
    \multicolumn{3}{@{}>{\raggedright}p{12.9cm}}{SIMS~\cite{speidel2023endovis}, SISVSE~\cite{yoon2022surgical}, Kvasir-Instrument~\cite{jha2021kvasir}, Dresden~\cite{carstens2023dresden}, DesmokeData~\cite{xia2024new}, LapGyn4~\cite{leibetseder2018lapgyn4}, Sinus Surgery~\cite{qin2020towards} (images); SPPIN~\cite{buser2025community} (MRI volumes)}
      & No video \\
    \multicolumn{3}{@{}>{\raggedright}p{12.9cm}}{EndoVis17~\cite{allan2019robotic} (temporally sampled frames), HiSWA-RLLS~\cite{speidel2024endovis} (5\,Hz), ROBUST-MIS~\cite{ross2021comparative} (one labeled frame per 10-s snippet)}
      & Sparse annotations or temporally sampled frames \\
    \multicolumn{3}{@{}>{\raggedright}p{12.9cm}}{DeSmoke-LAP~\cite{pan2022desmoke} (images and 10 short video clips), EndoNeRF~\cite{wang2022neural} (six clips of 4--8\,s), CVS challenge~\cite{alapatt2025sages} (90-s segments), SurgToolLoc~\cite{zia2023surgical} (30-s training clips)}
      & Short clips or few videos \\
    \multicolumn{3}{@{}>{\raggedright}p{12.9cm}}{EndoVis18~\cite{allan2020robotic} (nephrectomy), PitVis-2023~\cite{das2025pitvis} (pituitary surgery), PSI-AVA~\cite{valderrama2022towards} (prostatectomy), ART-Net~\cite{hasan2021detection} (hysterectomy), LapEx~\cite{derathe2025lapex} (sleeve gastrectomy), EgoSurgery~\cite{fujii2024egosurgery} (egocentric open surgery), SurgRIPE~\cite{xu2025surgripe} (phantom background), SurgPose~\cite{wu2025surgpose} (ex vivo tool motion), YouTube mixed collection (many procedures)}
      & Other procedure or none \\
    \bottomrule
  \end{tabular}
\end{table*}

The survey considered 36 datasets and two YouTube collections, yielding five sources (\cref{tab:B-survey}).
Suitable entries and AutoLaparo, initially undecided, formed the candidate pool.
Annotation focused on cholecystectomy and its robot-assisted simulation and training, so pooled prostatectomy and hysterectomy datasets were excluded.
PhaKIR, the remaining shortlisted cholecystectomy source, was proposed after Gold annotation was completed.

\subsection{Annotation tool and track propagation}
\label{supp:B2}

\begin{figure*}[p]
  \centering
  \includegraphics[width=0.93\textwidth]{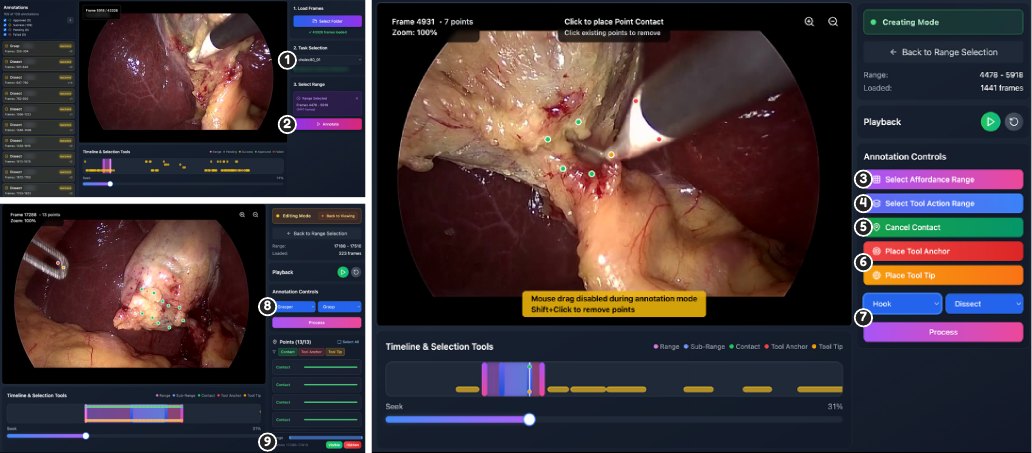}
  \caption{\textbf{Annotation interface and the nine-step workflow for constructing a tracked clip.} The panels show early interface examples.
  Top left: video overview with annotated clips and timeline.
  Right: local-range annotation, illustrated by a hook-dissect clip with seven points.
  The bottom-left panel shows a grasp contour in the earlier multi-point format. The final release and benchmark use four tracked tissue points for every action.
  Numbered steps:
  \step{1}~open the assigned video;
  \step{2}~select a local range;
  \step{3}~mark the action window;
  \step{4}~mark the interaction window;
  \step{5}~place contact points on one frame;
  \step{6}~place the tool anchor and tips, each on its own selected frame;
  \step{7}~assign tool and action classes and submit the tracking job;
  \step{8}~replay the tracks and revise the tool or action before approval;
  \step{9}~assign point visibility over a frame range.}
  \label{fig:B-tool}
\end{figure*}

Gold tracks each clip from sparse key-frame prompts (\cref{sec:construction}).

\paragraph{Annotation tool.}
The web tool in \cref{fig:B-tool} supports temporal labels and spatial prompts, with server-side tracking.
Reviewers inspect and revise clips in the same interface, and the release contains their latest annotations.

\paragraph{Propagation.}
The server uses offline CoTracker3~\cite{karaev2025cotracker3} with backward tracking, propagating each point in both temporal directions from its selected key frame.
Tracking covers the action window and all prompt frames; only frames within the window are retained.
Each retained frame stores the point coordinates in source-video pixels, visibility flags, and exact frame index.
Early grasp annotations used multiple points to outline the contact region. To simplify and standardize the annotation format across actions, these contours were converted to four tissue points, one at each corner of the minimum-area rectangle enclosing the annotated contact region.
The points were tracked again to obtain the final four tissue tracks used by the benchmark (\cref{supp:D2}). The standardized tracks were included in the final re-check.
Contact points enclose the affected tissue with a small buffer, following the tool-specific placement rules in \cref{supp:E1} and the tolerances in \cref{supp:B3}.

\subsection{Rules and tolerances of the two tiers}
\label{supp:B3}

\begin{table}[t]
  \centering
  \caption{\textbf{Annotation tolerances and review requirements for Gold and Silver.}
  The temporal tolerances for Gold and Silver follow the v1.0 and v2.0 annotation guides, respectively.
  Silver allows wider temporal tolerances and assigns spatial prompting to the generator. A tolerance is the maximum accepted deviation from the reference frame or position.}
  \label{tab:B-rules}
  \footnotesize
  \setlength{\tabcolsep}{3pt}
  \begin{tabular}{@{}>{\raggedright}p{1.45cm} >{\raggedright}p{3.05cm} >{\raggedright\arraybackslash}p{3.1cm}@{}}
    \toprule
    & Gold & Silver \\
    \midrule
    Tool, action   & \multicolumn{2}{l}{Exactly right} \\
    Missed actions & \multicolumn{2}{l}{Under 10\% of the actions} \\
    Action window ends & About ${\pm}5$ frames; whole action inside & ${\pm}10$--20 frames; whole action inside \\
    Interaction window ends & 1--2 frames; grasp end about 5 frames & 3--5 frames \\
    Tool points    & 5\,px; scissors pivot 2\,px; grasper tighter & Placed by the Silver generator \\
    Contact points & Enclose the contact region with a small buffer, allowing a placement tolerance of 5--10\,px & Placed by the Silver generator \\
    Review         & Two rounds, then reviewer correction; core-team re-check & One reviewer checks the human labels \\
    \bottomrule
  \end{tabular}
\end{table}

The requirements in \cref{tab:B-rules} specify accepted precision and annotation procedures. \Cref{supp:E1,supp:E2} provide the remaining protocol rules.

\subsection{Review and re-check}
\label{supp:B4}

\begin{table}[t]
  \centering
  \caption{\textbf{Annotation-stage Gold review records by annotator group and source.}
  Corrected counts clips that remained unacceptable after the second round and were fixed by the reviewer; the recorded corrections occur in the recruited-annotator group.
  The core team labeled 18 Cholec80 videos and one CRCD video.
  The review log contains 100 more YouTube records than the final release (2,866 vs.\ 2,766 in \cref{tab:sources}).}
  \label{tab:B-review}
  \footnotesize
  \setlength{\tabcolsep}{4.5pt}
  \begin{tabular}{@{}l rr rr@{}}
    \toprule
    & & Approved & \multicolumn{2}{c}{Corrected} \\
    \cmidrule(lr){4-5}
    Labeled by & Videos & records & records & \% \\
    \midrule
    \multicolumn{5}{@{}l}{\textit{Recruited annotators}} \\
    \quad Cholec80 & 12 & 2,813        & 120 & 4.3  \\
    \quad HeiChole & 10 & 1,799        & 111 & 6.2  \\
    \quad CRCD     & 7  & 727          & 130 & 17.9 \\
    \quad SurgVU   & 32 & 4,281        & 205 & 4.8  \\
    \quad YouTube  & 21 & 2,866        & 87  & 3.0  \\
    \quad All      & 82 & 12,486       & 653 & 5.2  \\
    \midrule
    Core team      & 19 & 3,282        & 0   & 0.0  \\
    \bottomrule
  \end{tabular}
\end{table}

The final Gold release contains 15,668 clips (\cref{tab:sources}).
\Cref{tab:B-review} reports annotation-stage review records, including 12,486 records produced by recruited annotators, of which 653 required reviewer correction after two review rounds (5.2\%).

\paragraph{Review.}
Records revised by their annotators are included in the approved total of the review log.

Corrections are concentrated in a small subset of videos: the six videos with correction rates above 20\% contain 5.9\% of the records and account for 39\% of corrections.
One CRCD video has the highest rate, 78\% (80 of 102 clips), and contributes most of that source's corrections.

\paragraph{Re-check.}
The final re-check assigned each Gold video to one core-team member for inspection of all finalized clips.
It included the grasp clips after standardization to four tissue tracks (\cref{supp:B2}). Changes made during this re-check were not counted (\cref{tab:B-quality}).

\subsection{Annotation effort}
\label{supp:B5}

Separating human semantic and temporal labeling from automated spatial prompting reduces the manual effort required for Silver annotation.

\paragraph{People.}
Sixteen recruited students annotated 82 Gold videos, and four core-team members annotated the remaining 19.
Three core-team members reviewed the student annotations, and five conducted the final Gold re-check.
Students annotated Silver following \cref{supp:E2}.

\subsection{The Silver generator}
\label{supp:B6}

For model-assisted annotation, we use an early version of the \net Spatial Head, trained on data selected for spatial prompt generation. Human annotators provide the tool and action labels and temporal boundaries, while the trained model generates spatial prompts for propagation.

\paragraph{Model.}
All four components follow the Spatial Head architecture of \net (\cref{supp:C1}).
They comprise the full-frame stage 1 that also defines the zoom region, the zoom stage for tool points, the contact decoder, and the refiner.
Full-frame and zoomed inputs are encoded by a frozen V-JEPA~2.1 ViT-L at 384 pixels. The refiner operates on native-frame crops.

\paragraph{Training data.}
The following describes the training data used for this annotation task. The benchmark experiments in the main paper follow the video-level partitions in \cref{supp:D1}.
For spatial prompt generation, the three tool-point components, stage 1, zoom, and refinement, were trained only on SurgVU.
An earlier split divided each of the 32 SurgVU Gold videos temporally into training, validation, and test segments (70/15/15\%), separated by 30\,s gaps.
Excluding clips in these gaps left 3,528 clips.
The tool-point components of the early generator evaluated in \cref{supp:B7} were trained on the training and validation segments.
The contact decoder is trained on targets derived from annotated tool-tip trajectories (\cref{eq:hindsight}) and evaluated against human-annotated tissue regions. It was trained on these SurgVU clips and 5,082 clips from the other 69 Gold videos, covering every pair with hook dissect subsampled.
Thus, only the contact decoder was trained on Gold videos outside SurgVU.

\paragraph{Point placement.}
The Silver release contains 7,327 clips from SurgVU.
For spatial annotation, Track Propagation (\cref{supp:C1}) selects the prompt frames, and CoTracker3 propagates the seven predicted points using the Gold settings (\cref{supp:B2}).

\subsection{Gold review and model-assisted annotation evaluation}
\label{supp:B7}

\begin{table}[t]
  \centering
  \caption{\textbf{Scope of Gold review and model-based annotation evaluations.}
  Human review covers every Gold clip. Model-based evidence covers re-annotation with Gold-trained \net and held-out evaluations of early tool-point and contact-decoder variants.
  PCK@8: share of tool points within 8 pixels.
  $^\dagger$Tool-point results use models trained on the training and validation segments. Contact Dice uses the training-only development variant in \cref{supp:C3}, without the additional 5,082 clips.}
  \label{tab:B-quality}
  \footnotesize
  \setlength{\tabcolsep}{3pt}
  \begin{tabular}{@{}>{\raggedright}p{1.7cm} >{\raggedright}p{3.55cm} >{\raggedright\arraybackslash}p{2.35cm}@{}}
    \toprule
    Evidence & Result & Scope \\
    \midrule
    Review (Gold, recruited annotators) & \Cref{tab:B-review}; per video, median 2.6\% corrected, none in 24 of 82 videos & Clips that still failed after two rounds \\
    Re-check (all Gold) & Every clip re-checked & Changes not counted \\
    Reference model (Gold test set) & Direct How and Where predictions in \cref{tab:results}; propagation examples in \cref{fig:reannot} & Gold-only benchmark model \\
    Early-model evaluations (417 held-out SurgVU clips) & Tool points: 6.56\,px at 256\,px, PCK@8 0.799. Contact Dice: 0.23 against human regions$^\dagger$ & Test segments of the earlier within-video split, using separate tool-point and contact-decoder variants \\
    \bottomrule
  \end{tabular}
\end{table}

\Cref{tab:B-quality} combines Gold review records with model-based evaluations on Gold-labeled clips.
Every Gold clip underwent review and a final re-check.
We assess Silver-style re-annotation on the Gold test set using Gold-trained \net and separately evaluate early tool-point and contact-decoder variants on held-out SurgVU clips.
These early evaluations use the within-video SurgVU split, while the main benchmark uses the held-out videos defined in \cref{supp:D1}.
\Cref{supp:C3} provides the corresponding contact-supervision comparison and further Spatial Head development analyses.

\section{\net architecture and implementation}
\label{supp:C}

This section details the architecture, objectives, implementation, and design choices of \net (\cref{sec:model}).

\subsection{Architecture and losses}
\label{supp:C1}

\begin{figure*}[t]
  \centering
  \includegraphics[width=\textwidth]{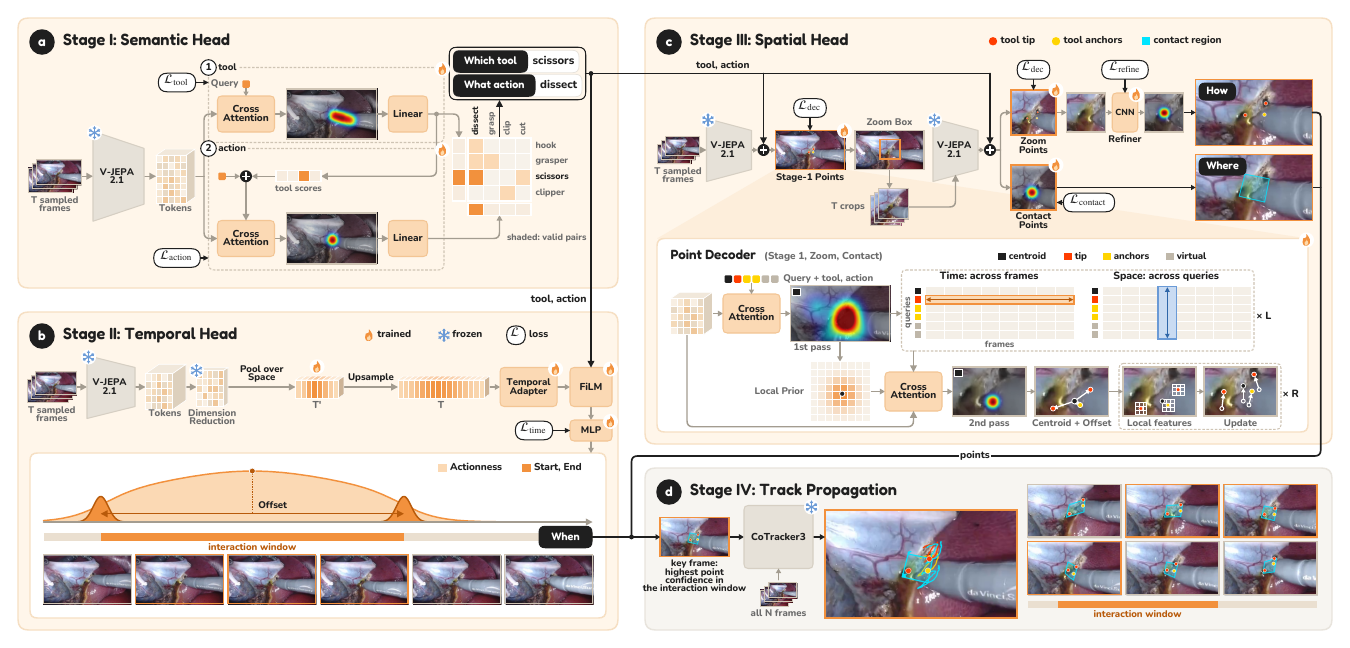}
  \caption{\textbf{Default \net architecture with prediction heads and dense track propagation.}
  (a) Semantic Head: tool and action queries cross-attend to frozen V-JEPA~2.1 tokens. The tool scores condition the action query, and the valid pairs (shaded) determine Which and What.
  (b) Temporal Head: a frozen block reduces the tokens; spatial pooling and upsampling produce one feature per frame. A temporal adapter, tool--action FiLM, and an MLP predict actionness, boundaries, and offsets to localize the interaction window (When).
  (c) Spatial Head: stage~1 predicts full-frame tool points and a zoom box for the tool and contact decoders. The refiner updates the zoomed tool points (How), while the contact points define the region (Where).
  In all three decoders, centroid, point, and virtual queries exchange information across frames and queries. A second pass adds a local prior, followed by $R$ updates.
  (d) Track Propagation: separately for tool points and corners, the most confident sampled frame inside the interaction window seeds CoTracker3, which propagates them to all $N$ frames.
  $\mathcal{L}$ denotes the losses in \cref{supp:C1}.
  Roman numerals I--IV identify the four architectural components. Within component III, stages 1 and 2 denote full-frame and zoomed tool-point prediction, respectively.}
  \label{fig:supp-model}
\end{figure*}

\Cref{fig:supp-model} shows four components: semantic, temporal, and spatial prediction from shared video tokens, followed by dense track propagation.
Their architectures and objectives follow below, with hyper-parameters in \cref{tab:supp-C-values}.

In the default configuration, the frozen encoder maps the $T$ sampled frames to tokens $\mathbf{Z}\in\mathbb{R}^{T'\times H'\times W'\times C}$, with each of the $T'$ tubelets covering $T/T'$ frames.

\begin{table*}[t]
  \centering
  \caption{\textbf{Architecture and optimization hyper-parameters of \net.}
  Symbols are defined in \cref{supp:C1}.}
  \label{tab:supp-C-values}
  \footnotesize
  \setlength{\tabcolsep}{4pt}
  \begin{tabular}{@{}l >{\raggedright\arraybackslash}p{\dimexpr\textwidth-2.9cm\relax}@{}}
    \toprule
    Part & Values \\
    \midrule
    Encoder & V-JEPA~2.1 ViT-L/16, frozen; input $384\times384$; $T=16$, $T'=8$, $H'=W'=24$, $C=1024$; full-frame last-layer tokens computed once per clip and cached in fp16 \\
    Semantic Head & One query per branch, of dimension $C$, with 16 attention heads; $\alpha$ starts at 0.05 \\
    Temporal Head & $C_{\mathrm{p}}=512$; $\mathbf{v}_t$ of dimension 256; adapter kernel 5; $\phi=0.5$; $K=15$ \\
    Temporal loss & $\lambda_{\mathrm{cls}}=0.1$, $\lambda_{\mathrm{b}}=0.1$, $\lambda_{\mathrm{prop}}=0.2$, $\lambda_{\mathrm{Q}}=0.15$, $\lambda_{\mathrm{ada}}=0.05$; focal loss with focusing parameter 1.5 and positive weight 0.25; sensitivity widths 0.5 (center) and 0.35 (ends) of half the interaction window, with floor 0.25; total weights 0.2 and 0.8 for the positive and the negative frames of a clip; varifocal loss with parameters 2 and 0.75 \\
    Point decoders & $P=3$ (tool) or 4 (contact); $V=8$; $L=2$; $R=4$; 16 attention heads in the cross-attention passes and query mixing, and 4 heads of width 128 in the update steps, whose sampled features are projected to 64 channels; median filter over 5 frames; soft-argmax window of $5\times5$ cells; offsets bounded by 0.2 and each update by 0.06, in normalized image coordinates; local features from $3\times3$ cells; $\psi$ starts at 0 \\
    Point loss & Huber threshold 6\,px; $\lambda_{\mathrm{ce}}=2$ with a Gaussian of 1 cell; $\lambda_{\mathrm{conf}}=0.5$ with an error threshold of 8\,px; both thresholds in pixels at $384\times384$ of the decoder's input (full frame or zoom crop). $(w_0,\dots,w_4)\propto(0.60, 0.50, 0.45, 0.40, 0.50)$ up to epoch 16, $(0.35, 0.45, 0.60, 0.80, 1.00)$ from epoch 20 to 44, and $(0.15, 0.25, 0.50, 0.90, 1.00)$ from epoch 50, linear in between \\
    Augmentation & Token masking rate 0 for 5 epochs, then linear to 0.075 at epoch 20 and to 0.15 at epoch 30, then constant. A masked token becomes a learned mask token with a position embedding (80\%) or a random token of its tubelet (10\%), or it stays (10\%). Centroid noise with probability 0.5 and standard deviation 0.015 in normalized coordinates \\
    Zoom box & Percentiles 5 and 95; $\mu=1.5$; $\nu=96$\,px; $B_{\min}=256$\,px; $B_{\max}=720$\,px (native pixels). Training boxes: center shifted by up to $0.08B$ per axis and side scaled by a factor in $[0.9, 1.25]$, drawn once per clip \\
    Contact targets & $\sigma_g=14W/384$ native px, isotropic in the frame ($W$: content-box width), \ie, 14\,px across and $14W/H$\,px down at $384\times384$; grid of $48\times48$ cells over the zoom box; $\eta=1.5$; $\lambda_{\mathrm{dice}}=60$; $\tau=0.5$ \\
    Refiner & $D=128$ native px; 12 input channels; $32\times32$ heatmap (4 native px per cell); input Gaussians of 6 native px; target Gaussian of 1 cell; $\lambda_{1}=0.025$ per native pixel; $\lambda_{\mathrm{tr}}=0.2$; 32 base channels \\
    \bottomrule
  \end{tabular}
\end{table*}

\paragraph{Semantic Head.}
The tool and action queries each attend to all $T'H'W'$ tokens through separate multi-head cross-attention layers.
A linear classifier maps each pooled feature to logits $\boldsymbol{\ell}^{\mathrm{t}}$ or $\boldsymbol{\ell}^{\mathrm{a}}$.
To incorporate the tool--action relationships in \cref{sec:definition}, the action query $\mathbf{q}^{\mathrm{a}}$ also receives the tool prediction as a prior.
This prior is initialized to zero and uses a stopped gradient to preserve the tool branch during action training:
\begin{equation}
  \mathbf{q}'^{\mathrm{a}} = \mathbf{q}^{\mathrm{a}} + \alpha\,\mathbf{W}\,\operatorname{sg}\big(\operatorname{softmax}(\boldsymbol{\ell}^{\mathrm{t}})\big),
  \label{eq:supp-toolprior}
\end{equation}
where $\mathbf{W}$ is a zero-initialized linear map, $\alpha$ is a learnable scalar, and $\operatorname{sg}$ denotes stop-gradient.
The two sets of logits are combined through \cref{eq:pair} to select a valid pair.

\paragraph{Temporal Head.}
Frame-level prediction requires converting tubelet tokens into one feature per frame.
A frozen channel-reduction block first maps $C$ channels to $C_{\mathrm{p}}$.
It retains its random initialization and comprises a linear projection, depthwise and pointwise convolutions, and a residual that moves each tubelet halfway toward the mean over tubelets.
Concatenated spatial mean and maximum features are mapped to $C_{\mathrm{p}}$ channels by a trained layer (LayerNorm, linear map, GELU), then linearly interpolated to the $T$ frames.
To adapt these frozen features for localization, a temporal adapter adds a zero-initialized residual branch, inspired by SoLa~\cite{kang2023soft}.
The pair also provides a cue for When (\cref{sec:stats}). Accordingly, FiLM~\cite{perez2018film} conditions each adapted feature $\mathbf{h}_t$ on $\boldsymbol{\kappa}$, the sum of a tool and an action embedding:
\begin{equation}
  \mathbf{h}'_t = \mathbf{h}_t \odot \big(1+\phi\tanh(\mathbf{W}_1\boldsymbol{\kappa})\big) + \mathbf{W}_2\boldsymbol{\kappa},
  \label{eq:supp-film}
\end{equation}
where $\mathbf{W}_1$ and $\mathbf{W}_2$ are linear maps and $\phi$ bounds the scale.
A shared MLP maps $\mathbf{h}'_t$ to $\mathbf{v}_t$, and linear layers predict the actionness logit $A_t$ and boundary logits $b^{\mathrm{s}}_t$ and $b^{\mathrm{e}}_t$.

Following TriDet~\cite{shi2023tridet}, each distance to an interval boundary is represented by a distribution over bins $d\in\{0,\dots,K\}$. Each bin incorporates the boundary logit of the frame at that distance:
\begin{equation}
  \zeta^{\mathrm{s}}_{t,d} = \beta^{\mathrm{s}}_{t,d} + b^{\mathrm{s}}_{t-d}, \qquad
  \zeta^{\mathrm{e}}_{t,d} = \beta^{\mathrm{e}}_{t,d} + b^{\mathrm{e}}_{t+d},
  \label{eq:supp-bins}
\end{equation}
where $\beta^{\mathrm{s}}$ and $\beta^{\mathrm{e}}$ are predicted from $\mathbf{v}_t$, and frames outside the action window contribute zero.
This formulation increases the weight of distances that align with likely boundaries.
The means of the two softmax distributions over $d$ yield $\hat{l}_t$ and $\hat{r}_t$ in \cref{eq:interval}.
Inspired by GFLV2's distribution-guided quality prediction~\cite{li2021generalized}, the quality logit $Q_t$ evaluates these distances using $\mathbf{v}_t$ and the top-1 probability, top-1 margin, and entropy of each distribution.
Finally, a proposal branch~\cite{lin2019bmn} regularizes $\mathbf{v}_t$ during training by scoring every candidate interval from its boundary features, mean feature, and length.

\paragraph{Point-decoder architecture.}
The stage-1, zoom, and contact decoders share an architecture of two cross-attention passes followed by $R$ updates.
For frame $t$, each decoder uses $1+P+V$ queries indexed by $n$: a centroid query, $P$ point queries, and $V$ virtual queries for context.
Each query combines a shared learned state, condition $\boldsymbol{\kappa}$, a query-specific embedding, and an embedding of the frame within its tubelet.
In the first pass, queries cross-attend to their tubelet's tokens, with keys modulated by FiLM from $\boldsymbol{\kappa}$.
This produces query states and attention maps $\pi^{(1)}_{t,n}$ over the $H'\times W'$ cells.
To exchange motion information across frames, query mixing operates on these states.
It stacks $L$ blocks, each with one transformer layer along time and one across the queries within a frame.
The resulting zero-initialized residual is added to the second-pass queries, which reuse the cross-attention weights.

The second pass further incorporates a learned local prior around the first-pass estimates.
For point $p$, a windowed soft-argmax around the peak of $\pi^{(1)}_{t,p}$, followed by temporal median filtering, gives a reference $\bar{\mathbf{y}}_{t,p}$.
The local prior added to the second-pass attention logits is
\begin{equation}
  a_{t,p}(\mathbf{g}) = -\,\omega_{t,p}\,\big\lVert(\mathbf{g}-\bar{\mathbf{y}}_{t,p})\odot(W',H')\big\rVert^2,
  \label{eq:supp-localprior}
\end{equation}
where $\mathbf{g}$ is the normalized center of a cell.
A zero-initialized MLP predicts $\omega_{t,p}$ from the peak value and normalized entropy of $\pi^{(1)}_{t,p}$ and the tool identity, controlling the strength of the prior.
The centroid query averages the point references and weights; virtual queries receive no local prior.

To represent shared tool motion and relative geometry, the decoder separates the centroid from the point layout.
The centroid $\mathbf{c}_t$ is obtained from the second-pass centroid map $\pi^{(2)}_{t,n}$ by soft-argmax, followed by a correction of at most one cell.
Each point state predicts a bounded offset $\mathbf{o}_{t,p}$, with zero mean over points.
The point's own map supplies a complementary estimate through its windowed soft-argmax $\hat{\mathbf{y}}^{(2)}_{t,p}$. The initial prediction blends the centroid-offset and map-based estimates:
\begin{equation}
  \mathbf{x}^0_{t,p} = (1-\gamma)\,(\mathbf{c}_t+\mathbf{o}_{t,p}) + \gamma\,\hat{\mathbf{y}}^{(2)}_{t,p},
  \label{eq:supp-init}
\end{equation}
where $\gamma=\sigma(\psi)$ with a learned scalar $\psi$.
Next, $R$ update steps refine the points using shared weights and a learned step embedding, as in CoTracker~\cite{karaev2024cotracker}.
At step $r$, token features sampled around the points and centroid update $\mathbf{c}^r$ through a state carried across steps.
Attention within each frame and along time then predicts a bounded update $\Delta\mathbf{o}^r$.
The new offsets are $\mathbf{x}^{r-1}-\mathbf{c}^r+\Delta\mathbf{o}^r$ minus their mean; adding them to $\mathbf{c}^r$ gives $\mathbf{x}^r$, with $\mathbf{x}^R$ as the final output.

\begin{samepage}
For subsequent Track Propagation, a small MLP predicts each point's confidence from its second-pass state and the peak value and entropy of $\pi^{(2)}_{t,p}$.
\par
\end{samepage}

\paragraph{Coarse-to-fine spatial refinement.}
To localize the tool at finer resolution, the stage-1 points from all $T$ frames define one square zoom box per clip in native pixels.
With $\Delta x$ and $\Delta y$ the ranges between a lower and an upper percentile of the points, the side of the box is
\begin{equation}
  B = \operatorname{clip}\big(\mu\max(\Delta x,\Delta y)+\nu,\;B_{\min},\;B_{\max}\big),
  \label{eq:supp-box}
\end{equation}
where $\mu$ and $\nu$ provide a margin and $B_{\min}$ and $B_{\max}$ bound the box size.
The box is centered on the middle of the two ranges. Areas outside the frame are zero-padded, and the $T$ crops are resized to the encoder input.
The refiner then uses a $D\times D$ native-frame crop around each zoomed tool point to refine its position at the original resolution.
Its input comprises the image, Gaussian maps of the tool points with the current point first, and one-hot maps of the tool and point role (tip or anchor).
A small residual encoder--decoder predicts a crop heatmap, whose full-map soft-argmax gives the refined point.
Refinement applies to tool landmarks; contact corners define a region rather than visible landmarks and retain their decoder predictions.

\paragraph{Track Propagation.}
For dense annotation, Track Propagation extends the sampled-frame predictions across the clip.
Among sampled frames inside the interaction window, it selects the frame with the highest mean confidence of the three tool points or the four corners, separately for each group.
If the interaction window contains no sampled frame, it uses the nearest sampled frame as a fallback.
CoTracker3 then propagates all tool and contact points using the settings in \cref{supp:B2}.

\paragraph{Semantic and temporal objectives.}
The prediction heads use separate training objectives. Because the action branch incorporates the tool prediction, the tool branch is trained first with $\mathcal{L}_{\mathrm{tool}}=\operatorname{CE}(\boldsymbol{\ell}^{\mathrm{t}},y^{\mathrm{t}})$, where $y^{\mathrm{t}}$ is the true tool.
The tool branch is then frozen, and its attention weights initialize the action attention. The action branch and tool prior are trained with $\mathcal{L}_{\mathrm{action}}=\operatorname{CE}(\boldsymbol{\ell}^{\mathrm{a}},y^{\mathrm{a}})$ for the true action $y^{\mathrm{a}}$.
For the Temporal Head, the frames inside the true interaction window $[s^*,e^*]$ are positive, and their distances to its ends are the targets.
The temporal objective is
\begin{equation}
  \begin{split}
  \mathcal{L}_{\mathrm{time}} &= \lambda_{\mathrm{cls}}\mathcal{L}_{\mathrm{cls}} + \mathcal{L}_{\mathrm{reg}}\\
  &\quad {}+ \lambda_{\mathrm{b}}\bigl(|\hat{s}-s^*|+|\hat{e}-e^*|\bigr)\\
  &\quad {}+ \lambda_{\mathrm{prop}}\mathcal{L}_{\mathrm{prop}} + \lambda_{\mathrm{Q}}\mathcal{L}_{\mathrm{Q}} + \lambda_{\mathrm{ada}}\mathcal{L}_{\mathrm{ada}}.
  \end{split}
  \label{eq:supp-time}
\end{equation}
Here $\mathcal{L}_{\mathrm{cls}}$ applies focal loss~\cite{lin2017focal} to $A_t$, with frame sensitivity peaking at the interaction-window center~\cite{shao2023action} and count-based balancing of positive and negative frames~\cite{liu2021tsi}.
The complementary regression term $\mathcal{L}_{\mathrm{reg}}$ applies DIoU loss~\cite{zheng2020distance} to the positive-frame distances, with sensitivity peaking at the interaction-window ends.
The $\lambda_{\mathrm{b}}$ term directly supervises the decoded boundaries by comparing $\hat{s}$ and $\hat{e}$ in \cref{eq:interval} with the true ends.
Two additional terms supervise interaction-window quality using tIoU.
$\mathcal{L}_{\mathrm{prop}}$ is the binary cross-entropy between proposal scores and their tIoU with $[s^*,e^*]$.
$\mathcal{L}_{\mathrm{Q}}$ applies varifocal loss~\cite{zhang2021varifocalnet} to $Q_t$, with the tIoU of the interaction window predicted at $t$ as the positive-frame target and zero elsewhere.
Finally, $\mathcal{L}_{\mathrm{ada}}$ provides temporal feature regularization following SoLa.
For distinct sampled frames $t\ne t'$, let $\mathbf{z}_t$ denote a linear projection of the adapted feature. Binary cross-entropy brings $\big(1+\cos(\mathbf{z}_t,\mathbf{z}_{t'})\big)/2$ toward $\sigma\big(16/(t-t')^2\big)$, which decreases with frame separation.

\paragraph{Point-decoder objectives.}
Each point decoder minimizes
\begin{equation}
  \mathcal{L}_{\mathrm{dec}} = \sum_{r=0}^{R} w_r\,\mathcal{L}^{r}_{\mathrm{pt}} + \lambda_{\mathrm{ce}}\mathcal{L}_{\mathrm{ce}} + \lambda_{\mathrm{conf}}\mathcal{L}_{\mathrm{conf}},
  \label{eq:supp-dec}
\end{equation}
where $\mathcal{L}^{r}_{\mathrm{pt}}$ applies Huber loss to visible points after update $r$, with $r=0$ corresponding to \cref{eq:supp-init}.
The weights $w_r$ sum to one and progressively emphasize later updates during training.
$\mathcal{L}_{\mathrm{ce}}$ directly supervises the second-pass maps through cross-entropy between $\pi^{(2)}$ and a Gaussian around the true point.
$\mathcal{L}_{\mathrm{conf}}$ uses binary cross-entropy to identify final points that are visible and lie within a fixed error threshold.
To approximate stage-1 localization errors, the zoom and contact decoders train on randomly shifted and scaled boxes around visible true points.
Training additionally masks a subset of input tokens and perturbs the centroid before the first update.

\paragraph{Tip-derived contact supervision.}
The contact decoder represents its tip-derived training target in two complementary forms: a region map and four corner points.
\Cref{eq:hindsight} defines the map $M_t$ on a grid over the zoom box.
The set $\mathcal{F}_t$ uses frames within the interaction window at the native frame rate, retaining those at or after the source frame corresponding to $t$ with a visible tip. If this set is empty, all frames with a visible tip in the interaction window are used.
Interaction windows without visible tips receive no contact target.
For the corner representation, let $\bar{\mathbf{u}}_t$ be the mean of the tip positions $\mathbf{u}_i$ over $\mathcal{F}_t$, and let $(\xi_{t,j},\mathbf{e}_{t,j})$, $j\in\{1,2\}$, be the eigenpairs of their covariance plus $\sigma_g^2\mathbf{I}$. A single position has zero covariance.
The four corner targets lie at the ends of the two principal axes, scaled by $\eta$:
\begin{equation}
  \mathbf{x}^*_{t,j\pm} = \bar{\mathbf{u}}_t \pm \eta\sqrt{\xi_{t,j}}\,\mathbf{e}_{t,j}.
  \label{eq:supp-corners}
\end{equation}
The corners are expressed in zoom-box coordinates and clipped to $[0,1]$ on each axis.
Because their order is arbitrary, each clip uses the one of the $4!$ assignments with the lowest visibility-weighted L1 distance to the final predictions over the $T$ frames.

The region map contributes to $\mathcal{L}_{\mathrm{contact}}=\mathcal{L}_{\mathrm{dec}}+\lambda_{\mathrm{dice}}\mathcal{L}_{\mathrm{dice}}$, where $\mathcal{L}_{\mathrm{dice}}$ compares $M_t$ with a soft rasterization of the predicted quadrilateral.
For this rasterization, corners $\mathbf{p}_1,\dots,\mathbf{p}_4$ are sorted by angle around their mean, placing the interior on the positive side of each edge.
In grid units, a cell $\mathbf{g}$ receives $\prod_{m=1}^{4}\sigma\big((\mathbf{p}_{m+1}-\mathbf{p}_m)\times(\mathbf{g}-\mathbf{p}_m)/\tau\big)$, where $\times$ is the 2D cross product, $\mathbf{p}_5=\mathbf{p}_1$, and $\tau$ is a temperature.
The score therefore approaches one inside all four edges.
Both targets use later-frame annotations. Query mixing provides bidirectional context over the sampled frames.

\paragraph{Refiner objective.}
To model stage~1 uncertainty, refiner training crops are displaced from true points by stage-1 errors sampled from validation clips.
Its loss is \mbox{$\mathcal{L}_{\mathrm{refine}}=\mathcal{L}_{\mathrm{hm}}+\lambda_{1}\mathcal{L}_{1}+\lambda_{\mathrm{tr}}\mathcal{L}_{\mathrm{tr}}$}.
Here $\mathcal{L}_{\mathrm{hm}}$ is the cross-entropy between the heatmap and a Gaussian around the true point, and $\mathcal{L}_{1}$ is the L1 loss of the refined point; both are evaluated on crops containing the true point.
$\mathcal{L}_{\mathrm{tr}}$ trains an auxiliary head to predict whether a windowed soft-argmax is closer to the true point than the crop center.
The auxiliary head is used for checkpoint selection and omitted at inference (\cref{tab:supp-C-schedule}).

\subsection{Implementation details}
\label{supp:C2}

The default frozen-encoder configuration uses the component-specific schedules in \cref{tab:supp-C-schedule} and the shared preprocessing and optimization settings described below.
Each frame is first cropped to its video's content box, excluding black borders and static interface panels, and resized to $384\times384$ without preserving the aspect ratio. This defines the metric coordinates in \cref{sec:metrics} (\cref{supp:D2}).
The resulting image is normalized using the ImageNet mean and standard deviation.
All components use AdamW with a weight decay of 0.05 and are validated after every epoch.
Except for the refiner, they use bf16 mixed precision, a constant learning rate, gradient-norm clipping at 1.0, and seed 42.
To account for the long-tailed pair distribution (\cref{fig:overview}), training clips are sampled in proportion to the inverse square root of their pair's training count.
These weights are normalized by their mean over the six pairs and capped at 10.
We train the stage-1 decoder before the zoom and contact decoders and the refiner. The Semantic and Temporal Heads are trained independently.

\begin{table}[t]
  \centering
  \caption{\textbf{Component-specific training schedules and checkpoint selection.}
  Criterion: validation checkpoint metric (lowest loss or ADE, highest tIoU).
  EMA: weight moving-average decay; averaged weights serve validation and testing.
  $^{a}$Early stopping after 20 epochs without loss reduction, counted from epoch 5.
  $^{b}$Training resumes from the selected tool epoch and continues to epoch 73, with stopping as in $^{a}$.
  $^{c}$Early stopping after 2 epochs without a tIoU gain of 0.002, counted from epoch 3.
  $^{d}$Peak learning rate of a one-cycle schedule with 10\% warm-up; gradient norm clipped at 5.
  $^{e}$Tool-point ADE on crops around stage-1 predictions; the auxiliary head selects between the windowed soft-argmax and crop center.
  Batch 24 fits a 24\,GB GPU. LR: learning rate.}
  \label{tab:supp-C-schedule}
  \footnotesize
  \setlength{\tabcolsep}{3.5pt}
  \begin{tabular}{@{}l c c c c l@{}}
    \toprule
    Part & Epochs & \makecell{Batch\\(clips)} & LR & EMA & Criterion \\
    \midrule
    Semantic, tool branch   & $\le$200\rlap{$^{a}$} & 64 & $5{\times}10^{-5}$ & 0.999 & loss \\
    Semantic, action branch & $\le$73\rlap{$^{b}$}  & 64 & $5{\times}10^{-5}$ & 0.999 & loss \\
    Temporal Head           & $\le$80\rlap{$^{c}$}  & 64 & $5{\times}10^{-5}$ & --    & tIoU \\
    Each point decoder      & 80                    & 24 & $10^{-4}$          & 0.999 & loss \\
    Refiner                 & 20                    & 4  & $10^{-3}$\rlap{$^{d}$} & -- & ADE\rlap{$^{e}$} \\
    \bottomrule
  \end{tabular}
\end{table}

\subsection{Spatial Head design analysis}
\label{supp:C3}

The development variants in \cref{tab:supp-C-design} use the earlier within-video SurgVU split in \cref{supp:B6}, with 2,692 training, 419 validation, and 417 test clips.
Training and held-out segments share videos. The main results in \cref{tab:results} use the final video-level benchmark. \Cref{supp:B7} separately reports Gold-test re-annotation and held-out evaluations of early tool-point and contact-decoder variants.

For stage~1, five of the six individual changes reduce validation error, and their combination lowers it from 12.76 to 11.07 pixels.
FiLM on the keys increases the error by 0.11 pixels; it is retained in the architecture on which the subsequent variants are built.

The contact-supervision comparison evaluates both variants against human-annotated tissue regions.
Training on human regions gives a test Dice of 0.421, compared with 0.232 for the tip-derived targets used by \net.
For model-assisted annotation, however, the four contact points predicted on one selected frame initialize CoTracker3, which propagates them throughout the action window (\cref{supp:C1}).
The spatial coverage of these initial points therefore matters for the resulting annotation.
During development, we observed that some predictions from the decoder trained on human regions covered only part of the instrument's activity during the interaction.
This motivated tip-derived targets, which incorporate visible tool-tip positions from the current frame onward through the temporal aggregation in \cref{eq:hindsight}.
We retained this supervision in \net to encourage more complete initial contact-prompt coverage for subsequent propagation, despite its lower Dice against human regions in the development comparison.
As noted in the main paper, the Where comparison in \cref{tab:results} therefore reflects differences in both architecture and supervision.

\begin{table}[t]
  \centering
  \caption{\textbf{Spatial Head development comparisons on the earlier within-video SurgVU split.}
  The stage-1 and joint-decoder rows use early validation clips, and the spatial-stage and contact-target rows use early test clips.
  $L$ and $V$ denote query-mixing blocks and virtual queries (\cref{supp:C1}).
  Each + row adds its change to the preceding row; the last three stage-1 rows instead start from query mixing.
  ADE is measured in pixels at $384\times384$ over visible points; PCK@12 is the share of visible points within 12 pixels.
  Hull IoU and Dice compare the predicted corner hull with the human-annotated region, following \cref{supp:D2}.
  The human-target decoder also trains on 5,082 clips from the other sources; without these clips, its Dice against human regions is 0.419.
  \yes: included in \net; \no: excluded from \net.}
  \label{tab:supp-C-design}
  \footnotesize
  \setlength{\tabcolsep}{3pt}
  \begin{tabular}{@{}>{\raggedright\arraybackslash}p{4.1cm} r r c@{}}
    \toprule
    Change & Before & After & Kept \\
    \midrule
    \multicolumn{4}{@{}l}{\textit{Stages of the Spatial Head: test ADE\,/\,PCK@12}} \\
    + zoom stage & 11.82/0.720 & 10.67/0.772 & \yes \\
    + refiner    & 10.67/0.772 & 10.17/0.787 & \yes \\
    \addlinespace[2pt]
    \multicolumn{4}{@{}l}{\textit{Stage-1 decoder: validation ADE}} \\
    + per-point map ($\gamma$, $\mathcal{L}_{\mathrm{ce}}$) & 12.76 & 12.45 & \yes \\
    + FiLM on the keys          & 12.45 & 12.56 & \yes \\
    + confidence                & 12.56 & 12.23 & \yes \\
    + query mixing              & 12.23 & 11.86 & \yes \\
    + $L{=}2$ blocks, $V{=}8$   & 11.86 & 11.57 & \yes \\
    + local prior               & 11.86 & 11.67 & \yes \\
    + both of the last two      & 11.86 & 11.07 & \yes \\
    \addlinespace[2pt]
    \multicolumn{4}{@{}l}{\textit{Contact decoder}} \\
    One joint tool--contact decoder & & & \no \\
    \quad validation tool ADE        & 10.08 & 11.49 & \\
    \quad validation hull IoU        & 0.264 & 0.263 & \\
    Tip-derived instead of human targets & & & \yes \\
    \quad test Dice to human regions    & 0.421 & 0.232 & \\
    \bottomrule
  \end{tabular}
\end{table}

\section{Evaluation protocol and additional analyses}
\label{supp:D}

This section specifies the benchmark partitions, model inputs, and task metrics for \cref{sec:experiments}.
\Cref{supp:D1,supp:D2} define the partitions and metrics; \cref{supp:D3,supp:D4} report results and computational cost of \net. \Cref{supp:D5,supp:D7,supp:D8,supp:D9} cover baseline adaptations, model comparisons, Silver-style re-annotation, and qualitative predictions.

\subsection{Video partitions and model inputs}
\label{supp:D1}

The Gold release contains 15,668 clips (\cref{tab:sources}). The benchmark uses a fixed set of 15,499 of these clips, split by video into 12,372 training, 1,604 validation, and 1,523 test clips (\cref{tab:supp-D-splits}). Every split covers all five sources and six pairs.
Whole videos are assigned while balancing source composition, action pairs, and clip duration (\cref{tab:supp-D-splits}).
The validation and test splits each target 10\% of the merged annotated frames in Gold.
Merged annotated frames are the union of a video's action windows, so overlapping windows contribute only once.
Gold and Silver contain 1,816,646 and 1,082,230 such frames, respectively, totaling 2,898,876 frames, or approximately 2.9 million.
Each held-out split contains three videos each from Cholec80, SurgVU, and YouTube, and one each from HeiChole and CRCD.
YouTube videos are grouped by channel before assignment, keeping each channel within one split.
Under these constraints, simulated annealing (30 restarts of 2,000 steps) selects the video set $\mathcal{V}$ from the candidate videos $\mathcal{P}$ by minimizing
\begin{equation}
  \mathcal{L}(\mathcal{V}) = 100\sum_{k}[3-n_k]_{+} + \sum_{x} w_x r_x^{2} + 20\sum_{k}\Delta_k^{2},
  \label{eq:supp-D-split}
\end{equation}
with $[a]_{+}=\max(0,a)$.

In \cref{eq:supp-D-split}, $k$ indexes pairs and $s$ sources. The count $n_k$ is the number of videos in $\mathcal{V}$ containing pair $k$; the first term encourages each pair to occur in at least three videos per split.
The second term weighs the squared deviation from the target share $\rho$ of every source $s$, pair $k$, and source--pair cell $sk$, with $r_x = \bigl(F_x(\mathcal{V})/F_x(\mathcal{P})-\rho\bigr)/\rho$.
$F_s$ is the merged frame count of source $s$, and $F_k$ and $F_{sk}$ sum the clip frames of pair $k$ (within source $s$).
The weights are $w_s=w_k=4$ and $w_{sk}=\min\bigl(1,\,F_{sk}(\mathcal{P})/30{,}000\bigr)$, so a rare cell weighs less.
The final term controls duration through $\Delta_k=\bar{\ell}_k(\mathcal{V})-\bar{\ell}_k(\mathcal{P})$, where $\bar{\ell}_k$ is the mean log duration in seconds of clips of pair $k$. It penalizes differences in this mean relative to the candidate set.
The test split is selected first with $\rho=0.1$. The validation split is then drawn from the remaining videos with $\rho=1/9$, again targeting 10\% of Gold.

\begin{table}[t]
  \centering
  \caption{\textbf{Gold clips used by the benchmark.} Each video belongs to one split.}
  \label{tab:supp-D-splits}
  \footnotesize
  \setlength{\tabcolsep}{8pt}
  \begin{tabular}{@{}lrr@{}}
    \toprule
    Split & Videos & Clips \\
    \midrule
    Training   & 79  & 12,372 \\
    Validation & 11  & 1,604 \\
    Test       & 11  & 1,523 \\
    \midrule
    Total      & 101 & 15,499 \\
    \bottomrule
  \end{tabular}
\end{table}

\paragraph{Clip inputs.}
Each clip holds 16 sampled frames of its action window, cropped to the video's content box (\cref{supp:D2}).
Each Gold benchmark clip contains at least one window frame with all three tool points annotated, regardless of visibility.
A sampled frame without all three points receives their coordinates from the previous sampled frame, with the points marked invisible.
For the first sampled frame, the fallback is initialized from the earliest window frame with all three tool points annotated.
The Gold-only, Silver-only, and Gold + Silver models are evaluated on the same Gold test set.

\subsection{Task-specific metrics and paired bootstrap}
\label{supp:D2}

The evaluation in \cref{sec:metrics} measures each question separately. The definitions below specify the scoring and aggregation used in the implementation.
All spatial metrics use one fixed $384\times384$ coordinate system per video.

\paragraph{Which, What, and Pair.}
Each accuracy is the share of the 1,523 test clips whose prediction equals the label.
For \net, the pair follows \cref{eq:pair}, and \cref{tab:results} and \cref{tab:ablation}(a) read Which and What from this pair, as they do for the classifiers.
\Cref{tab:ablation}(c) instead reads them from the argmax of the tool and action logits.
Pair mAcc averages the accuracy within each true pair over the six pairs, giving each pair equal weight.

\paragraph{When.}
Temporal localization compares the predicted and annotated interaction windows on the normalized action-window axis.
Let $f_1,\dots,f_{16}$ be the frame indices of the sampled frames, so that $f_1$ and $f_{16}$ are the ends of the action window.
The annotated interaction window $[a_s,a_e]$ then maps to $s=(a_s-f_1)/(f_{16}-f_1)$ and $e=(a_e-f_1)/(f_{16}-f_1)$, and the predicted interaction window $(\hat s,\hat e)$ of \cref{eq:interval} is clipped to $[0,1]$ as well.
On this axis,
\begin{equation}
  \mathrm{tIoU} = \frac{\max\bigl(0,\,\min(\hat e,e)-\max(\hat s,s)\bigr)}{\max(\hat e,e)-\min(\hat s,s)}
  \label{eq:supp-D-tiou}
\end{equation}
is averaged over the 1,523 clips.
R@$\theta$ is the share of clips with $\mathrm{tIoU}\geq\theta$, for $\theta\in\{0.3, 0.5, 0.7\}$.
The whole-window floor predicts $(\hat s,\hat e)=(0,1)$, so its tIoU equals the share of the action window that the interaction window fills.

\paragraph{How.}
The coordinate system starts from one fixed content box per video, which removes black borders and static interface panels.
The box is derived from the pixel motion in the middle frames of up to 40 clips of the video, so it uses no annotated points.
Points are normalized to this box and scaled by 384 in both axes.
Since the content box never exceeds the source frame, at most $1280\times792$ (\cref{tab:sources}), one metric pixel spans at most 3.3 source pixels horizontally and 2.1 vertically.
In these coordinates, let $\mathbf{p}_{tj}$ and $\hat{\mathbf{p}}_{tj}$ be the annotated and the predicted position of tool point $j$ in sampled frame $t$, and $v_{tj}\in\{0,1\}$ its visibility flag.
For \net, $\hat{\mathbf{p}}$ is the output of the refiner.
The error of clip $c$ is
\begin{equation}
  e_c = \frac{\sum_{t=1}^{16}\sum_{j=1}^{3} v_{tj}\,\lVert\hat{\mathbf{p}}_{tj}-\mathbf{p}_{tj}\rVert}{\sum_{t=1}^{16}\sum_{j=1}^{3} v_{tj}},
  \label{eq:supp-D-ade}
\end{equation}
and the ADE averages $e_c$ equally across clips with at least one visible tool point.
In contrast, the median, the share above 32\,px, and PCK@$\delta$ pool all visible tool points, so every point counts equally.
PCK@$\delta$ is the share of these points with an error below $\delta$ pixels, for $\delta\in\{4, 8, 12, 16\}$.

\paragraph{Where.}
Contact-region evaluation uses the four exported tissue tracks of each clip and their visibility flags across the sampled frames.
For a sampled frame $t$ with at least two visible annotated contact points, let $G_t$ be the convex hull of these points and $P_t$ the hull of the four predicted points; for SegFormer, $P_t$ is instead its predicted mask, with the logits upsampled to $384\times384$ and thresholded at zero.
Both hulls are rasterized in the $384\times384$ frame with a disc of radius 6\,px at each hull vertex, and two-point hulls also receive a 12-pixel-wide line. These widths apply to scoring. Annotation buffers follow \cref{supp:B2,supp:E1}.
Then
\begin{equation}
  \mathrm{Dice}_t = \frac{2\,|P_t\cap G_t|}{|P_t|+|G_t|},
  \label{eq:supp-D-dice}
\end{equation}
and the IoU uses the same regions.
Contact-point error measures location rather than overlap: the Hungarian algorithm matches the four predictions to the visible annotated points in each frame, and the matched distances are averaged.
The reported Where Dice averages $\mathrm{Dice}_t$ over all 21,070 scored test frames. Clips with more scored frames therefore receive greater weight.

\paragraph{Paired video bootstrap.}
For clip-averaged ADE in \cref{sec:silver}, we evaluate the Gold-only and Gold + Silver models on the same test clips and resample whole videos.
Let $d(c)$ be the difference in ADE between the two models on clip $c$, and $\mathcal{C}(v)$ the clips of video $v$.
Each of 2,000 resamples draws $V$ videos $v_1^{(b)},\dots,v_V^{(b)}$ with replacement from the $V$ test videos and computes
\begin{equation}
  \hat{\Delta}^{(b)} = \frac{\sum_{i=1}^{V}\sum_{c\in\mathcal{C}(v_i^{(b)})} d(c)}{\sum_{i=1}^{V}\bigl|\mathcal{C}(v_i^{(b)})\bigr|}.
  \label{eq:supp-D-boot}
\end{equation}
A video selected twice contributes its clips twice.
The 95\% confidence interval is bounded by the 2.5th and 97.5th percentiles of $\hat{\Delta}^{(b)}$. The point estimate is the mean of $d(c)$ over all clips.

\subsection{Results by threshold, pair, and source}
\label{supp:D3}

To characterize the aggregate results in \cref{tab:results}, we examine localization thresholds, action pairs, and source-specific performance.
\Cref{tab:supp-D-extra} reports threshold-based and complementary metrics for Gold-only and Gold + Silver training.
The whole-window reference reaches an R@0.3 of 72.4\% because the interaction window occupies at least 30\% of the action window in most clips.
The advantage of \net over this reference increases from 11 points at $\theta=0.3$ to 25 points at $\theta=0.7$, the threshold reported in \cref{tab:results}.

For the Gold-only model, averaging by clip gives an ADE of 19.15\,px (\cref{supp:D2}). Pooling visible point observations gives an ADE of 18.55\,px, with greater weight assigned to clips containing more visible points.
A quarter of the points lie within 4\,px, while the median is approximately half the ADE, indicating a heavy error tail.

Contact overlap also depends on annotation visibility: Where Dice increases from 0.096 with two visible points to 0.306 with four.
A two-point target forms a thin segment whose overlap is particularly sensitive to spatial displacement.

\begin{table}[t]
  \centering
  \caption{\textbf{Threshold-based evaluation and complementary spatial metrics.}
  All results use the Gold test set and show small aggregate changes after adding Silver.
  Gold denotes Gold-only training. Gold + Silver denotes joint training.
  Whole window: the floor of \cref{tab:results}.
  Every metric follows \cref{supp:D2}.
  Visible points: annotated contact points visible in the scored frame.}
  \label{tab:supp-D-extra}
  \footnotesize
  \setlength{\tabcolsep}{4pt}
  \begin{tabular}{@{}lccc@{}}
    \toprule
    Metric & Gold & Gold + Silver & Whole window \\
    \midrule
    \multicolumn{4}{@{}l}{\textit{When}} \\
    R@0.3 (\%)$\uparrow$ & 83.5 & 83.5 & 72.4 \\
    R@0.5 (\%)$\uparrow$ & 67.0 & 67.2 & 43.1 \\
    R@0.7 (\%)$\uparrow$ & 40.7 & 39.7 & 15.4 \\
    \midrule
    \multicolumn{4}{@{}l}{\textit{How}} \\
    ADE over points (px)$\downarrow$ & 18.55 & 18.37 & -- \\
    Median (px)$\downarrow$          & 9.9   & 9.7   & -- \\
    PCK@4 (\%)$\uparrow$             & 25.4  & 25.9  & -- \\
    PCK@8 (\%)$\uparrow$             & 44.8  & 44.8  & -- \\
    PCK@12 (\%)$\uparrow$            & 54.4  & 55.5  & -- \\
    PCK@16 (\%)$\uparrow$            & 61.0  & 62.1  & -- \\
    $>$32\,px (\%)$\downarrow$       & 20.3  & 19.5  & -- \\
    \midrule
    \multicolumn{4}{@{}l}{\textit{Where, human regions}} \\
    IoU$\uparrow$                          & 0.153 & 0.154 & -- \\
    Frames with Dice $\geq0.5$ (\%)$\uparrow$ & 12.6 & 12.4 & -- \\
    Dice, 2 visible points$\uparrow$       & 0.096 & 0.095 & -- \\
    Dice, 3 visible points$\uparrow$       & 0.185 & 0.184 & -- \\
    Dice, 4 visible points$\uparrow$       & 0.306 & 0.308 & -- \\
    Contact-point error (px)$\downarrow$   & 33.9  & 33.9  & -- \\
    \bottomrule
  \end{tabular}
\end{table}

\Cref{tab:supp-D-pair} evaluates When, How, and Where separately for each pair, showing that the most difficult pair depends on the task.
The changes from Silver are smaller than the differences between pairs: at most 0.020 in tIoU and 0.010 in human-region Dice for each pair.
R@0.7 varies more for rare pairs, by up to 9 points; these estimates are based on only 33 to 88 test clips per pair.

\begin{table*}[t]
  \centering
  \caption{\textbf{Task-dependent variation across action pairs.} Scissors cut is the most difficult pair for When, and grasper grasp for How and Where.
  Results of \net by pair on the Gold test set.
  G: the Gold-only model. G+S: the Gold + Silver model.
  Clips gives the number of Gold test clips for each pair. Metrics follow the task-specific scoring rules in \cref{supp:D2}.
  The How columns give the numbers behind the left panel of \cref{fig:pairsource}.}
  \label{tab:supp-D-pair}
  \footnotesize
  \setlength{\tabcolsep}{4.5pt}
  \begin{tabular}{@{}lr cccc cc cc@{}}
    \toprule
    & & \multicolumn{4}{c}{When} & \multicolumn{2}{c}{How} & \multicolumn{2}{c}{Where} \\
    \cmidrule(lr){3-6}\cmidrule(lr){7-8}\cmidrule(l){9-10}
    & & \multicolumn{2}{c}{tIoU$\uparrow$} & \multicolumn{2}{c}{R@0.7 (\%)$\uparrow$} & \multicolumn{2}{c}{ADE (px)$\downarrow$} & \multicolumn{2}{c}{Dice, human regions$\uparrow$} \\
    \cmidrule(lr){3-4}\cmidrule(lr){5-6}\cmidrule(lr){7-8}\cmidrule(l){9-10}
    Pair & Clips & G & G+S & G & G+S & G & G+S & G & G+S \\
    \midrule
    hook dissect     & 1,041 & 0.576 & 0.573 & 40.3 & 39.2 & 15.98 & 15.33 & 0.241 & 0.244 \\
    grasper grasp    &   258 & 0.594 & 0.592 & 34.9 & 37.2 & 33.04 & 33.92 & 0.217 & 0.218 \\
    grasper dissect  &    88 & 0.717 & 0.698 & 63.6 & 54.5 & 17.47 & 17.54 & 0.268 & 0.263 \\
    scissors dissect &    46 & 0.747 & 0.735 & 65.2 & 65.2 & 17.26 & 17.76 & 0.302 & 0.292 \\
    clipper clip     &    57 & 0.582 & 0.562 & 33.3 & 28.1 & 18.79 & 21.47 & 0.276 & 0.275 \\
    scissors cut     &    33 & 0.530 & 0.533 & 15.2 & 21.2 & 18.24 & 16.23 & 0.266 & 0.257 \\
    \bottomrule
  \end{tabular}
\end{table*}

\Cref{tab:supp-D-source} compares tool-point localization across sources. YouTube has the highest error under Gold-only and Gold + Silver training.
HeiChole and CRCD each have one test video.

\begin{table}[t]
  \centering
  \caption{\textbf{Tool-point localization by source, with the highest errors on YouTube.}
  Tool-point ADE (px) of \net by source, the numbers behind the right panel of \cref{fig:pairsource}.
  Vid.: Gold test videos.
  Change: Gold + Silver minus Gold.}
  \label{tab:supp-D-source}
  \footnotesize
  \setlength{\tabcolsep}{4pt}
  \begin{tabular}{@{}lrccc@{}}
    \toprule
    Source & Vid. & Gold & Gold + Silver & Change \\
    \midrule
    Cholec80 & 3 & 15.99 & 16.34 & $+0.35$ \\
    SurgVU   & 3 & 19.73 & 18.52 & $-1.21$ \\
    YouTube  & 3 & 26.79 & 26.77 & $-0.02$ \\
    HeiChole & 1 & 16.15 & 15.26 & $-0.89$ \\
    CRCD     & 1 & 18.10 & 19.34 & $+1.24$ \\
    \bottomrule
  \end{tabular}
\end{table}

\paragraph{Training with Silver alone.}
To further assess Silver as a training resource, we train all prediction heads on Silver alone, using the same frozen video encoder, training recipes, and Gold validation and test sets as the Gold and Gold + Silver settings.
The Gold validation set supplies checkpoint selection and the refiner's error bank under the existing recipe (\cref{supp:C1,supp:C2}).
Silver provides human class and temporal labels together with model-generated spatial annotations (\cref{supp:B6}).
Gold and Silver use disjoint sets of video IDs.
\Cref{tab:ablation}(c) reports the resulting recognition, temporal, and spatial scores under the common evaluation protocol.
For Gold + Silver, tIoU decreases from 0.592 to 0.587, Dice increases from 0.242 to 0.244, and ADE decreases by 0.22 pixels, with a 95\% confidence interval of $[-1.28,0.89]$.
Silver-only training achieves a human-region Dice of 0.271.

\Cref{tab:supp-D-silver-only} examines tool-point localization across test sources.
On SurgVU, Silver-only training gives an ADE of 18.67\,px, compared with 19.73\,px for Gold-only training and 18.52\,px for Gold + Silver.
On SurgVU, Silver-only training has a lower point estimate of ADE than Gold-only training, while the paired 95\% confidence interval for their difference includes zero.
Across the other four sources, the corresponding errors are 30.10, 18.94, and 19.07\,px.
The source-dependent results support combining Silver's additional supervision with Gold's broader coverage.

\begin{table*}[t]
  \centering
  \caption{\textbf{Tool-point localization with Gold, Silver, and their combination.}
  G, G+S, and S denote Gold-only, Gold + Silver, and Silver-only training.
  Values are clip-averaged ADE in the $384\times384$ coordinate system, with lower values indicating smaller errors.
  Differences use unrounded per-clip errors, with 95\% confidence intervals from 2,000 paired bootstrap resamples of test videos.
  Other sources are Cholec80, HeiChole, CRCD, and YouTube.}
  \label{tab:supp-D-silver-only}
  \footnotesize
  \setlength{\tabcolsep}{4pt}
  \begin{tabular}{@{}lrccccc@{}}
    \toprule
    Test subset & Videos & G & G+S & S & $\mathrm{S}-\mathrm{G}$ & $\mathrm{S}-(\mathrm{G+S})$ \\
    \midrule
    All sources & 11 & 19.15 & 18.93 & 27.10 & \makecell{$+7.96$\\$[2.23,14.86]$} & \makecell{$+8.18$\\$[2.99,14.58]$} \\
    SurgVU & 3 & 19.73 & 18.52 & 18.67 & \makecell{$-1.06$\\$[-2.76,2.66]$} & \makecell{$+0.16$\\$[-3.15,1.32]$} \\
    Other four sources & 8 & 18.94 & 19.07 & 30.10 & \makecell{$+11.15$\\$[5.23,18.19]$} & \makecell{$+11.02$\\$[5.33,17.83]$} \\
    \bottomrule
  \end{tabular}
\end{table*}

\subsection{Computational cost}
\label{supp:D4}

\Cref{tab:supp-D-compute} reports hardware and elapsed times for training and inference.
The default frozen encoder supports training on cached tokens.
Two GPUs process separate module groups in parallel: one trains stage~1, the zoom stage, and the refiner sequentially, while the other trains the Semantic Head, Temporal Head, and contact decoder.

\begin{table}[t]
  \centering
  \caption{\textbf{Hardware and wall-clock cost of \net.}
  Export: 16 frames, content crop, and labels of every clip.
  Token cache: V-JEPA~2.1 tokens of every clip in fp16.
  Training: wall-clock time for all heads, including zoom-token computation and evaluation.
  Fine-tuning: the five runs in \cref{tab:ablation}(c), scheduled one per GPU at a time, with frame encoding at every step.
  Inference: both encoder passes and every Spatial Head stage, excluding frame decoding.}
  \label{tab:supp-D-compute}
  \footnotesize
  \setlength{\tabcolsep}{4pt}
  \begin{tabular}{@{}lll@{}}
    \toprule
    Step & Hardware & Time \\
    \midrule
    Clip export                   & 32 vCPUs                   & 1.2\,h \\
    Token cache                   & 2$\times$RTX 4090          & 1.3\,h \\
    Training, Gold only           & 2$\times$RTX 4090          & 19\,h \\
    Training, Gold + Silver       & 2$\times$RTX 4090          & 30\,h \\
    Encoder fine-tuning, 5 runs   & 4$\times$H100               & 10\,h \\
    Inference, Spatial Head       & 1$\times$RTX 4090          & 0.83\,s/clip \\
    \bottomrule
  \end{tabular}
\end{table}

\subsection{Baseline recipes and task-specific adaptations}
\label{supp:D5}

\begin{table*}[t]
  \centering
  \caption{\textbf{Published baseline recipes and their adaptations to \tactus.}
  The final column lists method-specific adaptations in addition to the shared changes in \cref{supp:D5}; original recipe values appear in parentheses.
  Input: frames $\times$ image size (width $\times$ height), or time steps of the frozen features.
  Epochs $a{+}b$: warm-up and main epochs; it: iterations.
  Batch: clips, single frames (DARK, ViTPose, SegFormer), or key frames (DSTA) per optimizer step.
  LR: peak learning rate.
  $\dagger$: frozen.}
  \label{tab:supp-D5}
  \footnotesize
  \setlength{\tabcolsep}{3.2pt}
  \newcommand{\rec}[1]{\\[-1pt]{\scriptsize\color{black!55}#1}}
  \begin{tabular}{@{}l l l l c c c >{\raggedright\arraybackslash}p{4.7cm}@{}}
    \toprule
    Method & Backbone & Pretraining & Input & Epochs & Batch & LR & Changes for \tactus (official value) \\
    \midrule
    \multicolumn{8}{@{}l}{\textit{(a) Which and What: video classifiers fine-tuned with six pair outputs}} \\
    \makecell[tl]{TSN~\cite{wang2016tsn}\rec{TSM repository, SSv2}}        & ResNet-50 & ImageNet-1k    & $16{\times}224^2$ & 50 & 64 & 1e-2    & 16 frames (8) \\
    \makecell[tl]{TSM~\cite{lin2019tsm}\rec{TSM repository, fine-tuning}}  & ResNet-50 & SSv2           & $16{\times}224^2$ & 25 & 64 & 1e-3    & 16 frames as its checkpoint (recipe: 8) \\
    \makecell[tl]{VideoMAE~\cite{tong2022videomae}\rec{SSv2 fine-tuning}} & ViT-B/16 & VideoMAE, SSv2 & $16{\times}224^2$ & 30 & 64 & 1.25e-4 & batch 64 (384); horizontal flip on (off for SSv2 only) \\
    \makecell[tl]{X-CLIP~\cite{ni2022xclip}\rec{K400, 16 frames}}    & ViT-B/16  & CLIP, K400     & $16{\times}224^2$ & 30 & 64 & 2e-6    & one fixed prompt per pair; batch 64 (256) \\
    \makecell[tl]{Endo-FM~\cite{wang2023endofm}\rec{PolypDiag fine-tuning}} & ViT-B/16 & Endo-FM     & $16{\times}224^2$ & 20 & 64 & 2.5e-4  & 16 frames (8), time embeddings interpolated; batch 64 (4) \\
    \midrule
    \multicolumn{8}{@{}l}{\textit{(b) When: heads trained from scratch on the frozen features of \net}} \\
    \makecell[tl]{ActionFormer~\cite{zhang2022actionformer}\rec{THUMOS14}} & ViT-L/16$^\dagger$ & V-JEPA~2.1 & 8 steps  & 5+30  & 2  & 1e-4 & 3 pyramid levels (6); global attention (local window 19) \\
    \makecell[tl]{TriDet~\cite{shi2023tridet}\rec{THUMOS14}}         & ViT-L/16$^\dagger$ & V-JEPA~2.1 & 8 steps  & 20+20 & 2  & 1e-4 & 3 pyramid levels (6); 4 boundary bins (16) \\
    \makecell[tl]{QD-DETR~\cite{moon2023qddetr}\rec{QVHighlights}}   & ViT-L/16$^\dagger$ & V-JEPA~2.1 & 8 steps  & 200   & 32 & 1e-4 & a fixed sentence per pair as query; saliency labels from the interaction window; negative queries formed by rolling the batch by one sample; ends not rounded (rounded to 2-s clips) \\
    \makecell[tl]{MS-TCN~\cite{farha2019mstcn}\rec{released defaults}} & ViT-L/16$^\dagger$ & V-JEPA~2.1 & 32 steps & 50   & 1  & 5e-4 & per-pair sigmoid outputs (softmax); 6 layers per stage (10) \\
    \midrule
    \multicolumn{8}{@{}l}{\textit{(c) How and Where: fine-tuned end to end on the full content frame}} \\
    \makecell[tl]{DARK~\cite{zhang2020dark}\rec{COCO, Adam}}         & HRNet-W32 & ImageNet-1k      & $1{\times}384{\times}288$ & 210 & 112 & 1e-3 & input turned to landscape ($288{\times}384$) \\
    \makecell[tl]{ViTPose~\cite{xu2022vitpose}\rec{COCO, ViT-B}}     & ViT-B/16  & MAE, ImageNet-1k & $1{\times}256{\times}192$ & 210 & 512 & 5e-4 & input turned to landscape ($192{\times}256$) \\
    \makecell[tl]{DSTA~\cite{he2024dsta}\rec{PoseTrack17, ResNet-50}} & ResNet-50 & ImageNet-1k     & $5{\times}256{\times}192$ & 20  & 32  & 5e-5 & key frame $\pm2$ raw frames ($\pm1$); ImageNet backbone (COCO pose checkpoint); local joint groups: the 4 tools and 6 pairs (limbs) \\
    \makecell[tl]{SegFormer~\cite{xie2021segformer}\rec{ADE20K, 160k}} & MiT-B2  & ImageNet-1k      & $1{\times}512^2$ crop     & 160k\,it & 16 & 6e-5 & per-pair sigmoid maps with BCE + soft Dice (softmax CE); human contact region as target \\
    \bottomrule
  \end{tabular}
\end{table*}

The comparisons in \cref{sec:baselines} adapt published recipes to the common Gold training set as follows.
\Cref{tab:supp-D5} identifies the reference recipe for each learned baseline in \cref{tab:results}.
Beyond the listed adaptations, each baseline retains the reference architecture, input size, optimizer, learning-rate schedule, augmentation, loss, decoding, and number of epochs (iterations for SegFormer).
Each baseline is trained on one RTX 4090 GPU with 24\,GB; gradient accumulation preserves the reference batch size where needed.
VideoMAE, X-CLIP, and Endo-FM instead use batches of 64 clips, with the reference learning rate scaled linearly.
All baselines use the same 12,372 Gold training clips as \net, with no Silver data.
The selected checkpoint maximizes validation Pair accuracy, tIoU, or Dice, or minimizes ADE, according to the task. Reported runs use the published configuration with seed~0.
The Spatial Head of \net was developed through comparisons on an earlier SurgVU split (\cref{supp:C3}), separately from these baseline recipes.

\paragraph{Task outputs and class conditioning.}
Outputs are defined per action pair, with tool-point outputs defined per tool.
X-CLIP scores a fixed prompt for each pair, such as ``a video of a hook dissecting tissue''.
ActionFormer, TriDet, and MS-TCN use one output channel per pair. Training and inference select the channel for the true pair. For MS-TCN, the per-frame argmax readout is replaced by the contiguous run above 0.5 with the highest summed probability. Boundaries are interpolated at the 0.5 crossings and extended to an action-window boundary when the run reaches it; the whole action window is returned when no run exists.
QD-DETR supplies the pair as a fixed text query, such as ``a hook dissects tissue'', encoded by the CLIP ViT-B/32 text encoder used in its reference recipe.
The keypoint models provide 36 outputs: three tool points per tool, reflecting their tool-specific roles, and four contact points per pair.
Training supervises the seven outputs associated with the true tool and pair, weighted by visibility; evaluation reads the same seven outputs.
The four exported contact tracks have clip-dependent ordering.
Training therefore orders each clip's tracks by the angle of their mean positions. Hull-based Dice is invariant to this ordering.

\paragraph{Input construction and temporal sampling.}
For classification, each sampled content frame is resized to a square and augmented at the reference pre-crop size.
At test time, the complete frame at $224\times224$ replaces center cropping to retain tool tips near the image boundary.
Keypoint models also retain the complete frame, resized to the reference input dimensions in landscape orientation, without an object box or detector.
DSTA additionally uses the two raw frames on each side of a sampled frame. During each epoch, it selects 8 of the 16 sampled frames per clip as key frames.
The single-frame methods DARK, ViTPose, and SegFormer train on sampled frames nearest to each whole second of the action window. At test time, these models predict on all 16 sampled frames. Scoring follows the visibility rules in \cref{supp:D2}.
SegFormer training retains only frames with at least two visible contact points.
Temporal methods use the frozen V-JEPA~2.1 tokens of \net, spatially averaged over the $24\times24$ grid.
The temporal sequence contains 8 steps from 16 sampled frames; MS-TCN instead receives 32 steps from 64 frames sampled uniformly over the action window.
These short sequences motivate the reductions in pyramid levels, boundary bins, and layers listed in \cref{tab:supp-D5}.

\paragraph{Initialization and training.}
All How and Where baselines learn their task-specific points and regions from Gold, without initialization from pose, detection, or segmentation checkpoints.
The keypoint and contact models load ImageNet-1k or MAE backbone weights and initialize their decoders from scratch.
For DSTA, this backbone initialization replaces the COCO pose checkpoint in its released configuration.
Human-pose half-body cropping and flip testing are removed from the keypoint recipes.
Horizontal flips are also removed to preserve the ordering of mirrored points.
SegFormer likewise trains without flips, making this setting common to all How and Where baselines; classifiers retain their reference flipping augmentation.
Image and video models train with bf16 mixed precision. For single-GPU SegFormer training, BatchNorm replaces SyncBN in the decoder.
Validation is performed every 10 epochs for DARK and ViTPose, ten times over training for SegFormer, and after each epoch for the other baselines.

\paragraph{Non-parametric references.}
The majority-class reference predicts hook dissect, the most frequent training pair.
The pair-mean reference averages each pair's training start and end positions on the normalized action-window axis (\cref{supp:D2}), quantifying the temporal information available from the pair alone.

\subsection{Controlled model comparisons}
\label{supp:D7}

\begin{figure*}[t]
  \centering
  \includegraphics[width=\textwidth]{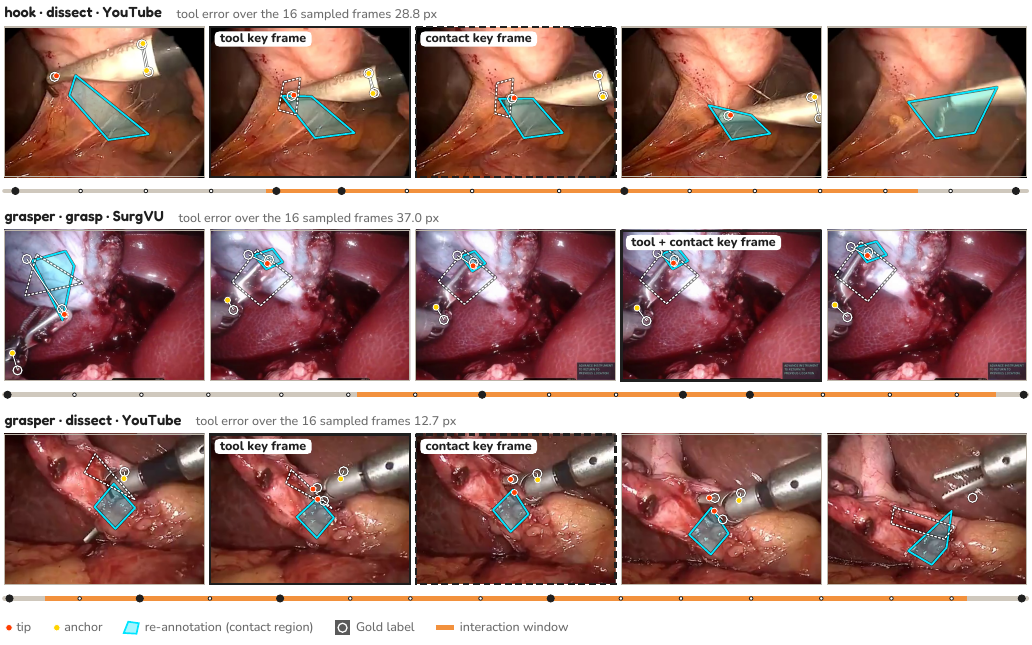}
  \caption{\textbf{Silver-style re-annotation on three random test clips, with tool-point errors of 12.7--37.0 pixels.}
  With the ground-truth tool, action, and interaction window supplied, Gold-trained \net selects its own prompt frames (solid box: tool; dashed box: contact), and CoTracker3 propagates the points throughout the action window.
  Filled dots and cyan regions: re-annotation. White circles and dashed lines: Gold.
  Each row shows 5 of the 16 sampled frames; the bar shows the action window (gray), the interaction window (orange), and the sampled frames.
  The clips are drawn at random, at most one per pair.}
  \label{fig:reannot}
\end{figure*}

We examine the backbone, spatial stages, and class conditioning of \net through \cref{tab:ablation}(a), (b), and (d).

\begin{table}[t]
  \centering
  \caption{\textbf{Frozen backbone configurations for \cref{tab:ablation}(a).} All four encoders provide tokens in the temporal layout used by \net.
  Input: frames $\times$ image size.
  Tokens: time steps $\times$ grid $\times$ width, where each step covers two consecutive sampled frames.
  $^*$: as in \cref{tab:results}.}
  \label{tab:supp-D7}
  \footnotesize
  \setlength{\tabcolsep}{5pt}
  \newcommand{\ck}[1]{\\[-1pt]{\scriptsize\color{black!55}#1}}
  \begin{tabular}{@{}l c c@{}}
    \toprule
    Backbone and checkpoint & Input & Tokens \\
    \midrule
    \makecell[tl]{VideoMAE ViT-L/16\ck{K400, self-supervised, no fine-tuning}}   & $16{\times}224^2$ & $8{\times}14^2{\times}1024$ \\
    \makecell[tl]{Endo-FM$^*$ ViT-B/16\ck{endoscopy videos, teacher weights}}     & $16{\times}224^2$ & $8{\times}14^2{\times}768$ \\
    \makecell[tl]{V-JEPA~2 ViT-L/16~\cite{assran2025vjepa2}\ck{public model at 256 pixels}}              & $16{\times}256^2$ & $8{\times}16^2{\times}1024$ \\
    \makecell[tl]{V-JEPA~2.1 ViT-L/16 (\net)~\cite{murlabadia2026vjepa}\ck{distilled from ViT-G, 384 pixels}} & $16{\times}384^2$ & $8{\times}24^2{\times}1024$ \\
    \bottomrule
  \end{tabular}
\end{table}

\paragraph{Backbones (\cref{tab:ablation}(a)).}
\Cref{tab:supp-D7} lists the four frozen encoders and their public checkpoints.
Each encoder processes the 16 sampled frames of a Gold clip once and produces the temporal layout of \net: 8 steps, each covering two consecutive frames, on its native token grid.
Preprocessing follows \net: frames are resized to the encoder's input size and normalized with ImageNet statistics. Endo-FM retains its own normalization (mean 0.45, standard deviation 0.225).
The VideoMAE and V-JEPA encoders produce these steps directly, because their tubelets span two frames.
Endo-FM produces one token per frame and spatial patch.
Its eight learned time embeddings are interpolated to the 16 frames, after which tokens from frames $2k$ and $2k{+}1$ are averaged into step $k$.
The heads adapt to the token grid and feature width, without an additional external layer.
The Semantic Head, Temporal Head, and stage 1 of the Spatial Head are retrained on each token set using the full recipe in \cref{supp:C2}; stage 1 again trains for 80 epochs with batches of 24 clips.
The V-JEPA~2.1 configuration corresponds to the default \net.

\paragraph{Spatial Head stages (\cref{tab:ablation}(b)).}
This comparison evaluates successive outputs of the same trained Spatial Head. A single run on each test clip records tool points after stage 1, zoom, and refinement; each output is scored directly without retraining.

\paragraph{Predicted pair (\cref{tab:ablation}(d)).}
This comparison isolates the effect of class-conditioning errors by supplying the same trained model with its predicted pair.
The Temporal Head is re-evaluated on all test clips using the predicted pair from \cref{eq:pair}.
The full Spatial Head, including stage 1, zoom-box construction, contact decoding, and refinement, is re-evaluated on clips with an incorrect pair prediction.
Clips with a correct pair prediction retain their protocol-run outputs.
The two rows of (d) therefore use identical trained heads and differ only in the supplied class condition.

\begin{table*}[t]
  \centering
  \caption{\textbf{Per-head encoder adaptation yields stronger classification and tool-point results than shared fine-tuning.} Shared-encoder changes relative to the frozen control remain small.
  All runs initialize from the trained heads of \net and continue for 10 epochs. Epoch 0 ($^\ddagger$) reports the adaptation evaluation run, separate from the main benchmark run in \cref{tab:results}.
  Epochs: the selected epoch, or the epochs for the Semantic, Temporal, and Spatial Head.
  Which and What are read from the predicted pair, as in \cref{tab:results}. \Cref{tab:ablation}(c) reports them from the separate head logits.
  Differences: Which, What, Pair, R@0.7, and $>$32\,px in points, tIoU and Dice $\times100$, ADE in pixels; 95\% CIs from the paired bootstrap over test videos below each value, bold where the CI excludes zero.}
  \label{tab:supp-ft}
  \scriptsize
  \setlength{\tabcolsep}{2.2pt}
  \begin{tabular}{@{}llcccccccccccc@{}}
    \toprule
    & & Which & What & \multicolumn{2}{c}{Pair} & \multicolumn{2}{c}{When} & \multicolumn{3}{c}{How} & Where & \multicolumn{2}{c}{How, by stage} \\
    \cmidrule(lr){3-3}\cmidrule(lr){4-4}\cmidrule(lr){5-6}\cmidrule(lr){7-8}\cmidrule(lr){9-11}\cmidrule(lr){12-12}\cmidrule(l){13-14}
    Encoder & Epochs & Acc & Acc & Acc & mAcc & tIoU & R@0.7 & ADE & Median & $>$32\,px & Dice & Stage 1 & Stage 2 \\
    \midrule
    frozen (\net)$^\ddagger$                & 0        & 96.3 & 96.5 & 93.8 & 82.7 & 0.591 & 40.5 & 19.15 & 9.86 & 20.3 & 0.242 & 20.24 & 19.42 \\
    frozen, +10 epochs                     & 1, 2, 1  & 96.3 & 96.8 & 94.0 & 83.0 & 0.591 & 40.6 & 19.08 & 9.81 & 20.3 & 0.242 & 20.29 & 19.34 \\
    fine-tuned, shared                     & 1        & 96.3 & 96.9 & 94.1 & 84.9 & 0.592 & 40.8 & 19.10 & 9.75 & 20.3 & 0.244 & 20.22 & 19.38 \\
    \quad epoch per group                 & 1, 4, 1  & 96.3 & 96.9 & 94.1 & 84.9 & 0.590 & 40.2 & 19.10 & 9.75 & 20.3 & 0.244 & 20.22 & 19.38 \\
    fine-tuned, per head                   & 10, 1, 6 & 97.6 & 97.8 & 95.8 & 89.2 & 0.592 & 40.4 & 18.68 & 9.29 & 19.9 & 0.244 & 19.78 & 18.84 \\
    \bottomrule
  \end{tabular}

  \vspace{6pt}
  \begin{tabular}{@{}lcccccccc@{}}
    \toprule
    Difference (first $-$ second) & Which & What & Pair & tIoU & R@0.7 & ADE & $>$32\,px & Dice \\
    \midrule
    shared $-$ per head & \makecell{\textbf{$-1.3$}\\{\scriptsize$[-2.4, -0.5]$}} & \makecell{\textbf{$-0.9$}\\{\scriptsize$[-1.7, -0.1]$}} & \makecell{\textbf{$-1.7$}\\{\scriptsize$[-3.1, -0.5]$}} & \makecell{$0.00$\\{\scriptsize$[-0.39, 0.33]$}} & \makecell{$+0.4$\\{\scriptsize$[-0.7, 1.7]$}} & \makecell{\textbf{$+0.42$}\\{\scriptsize$[0.18, 0.69]$}} & \makecell{$+0.42$\\{\scriptsize$[-0.07, 0.91]$}} & \makecell{$+0.06$\\{\scriptsize$[-0.20, 0.21]$}} \\
    shared (epoch per group) $-$ per head & \makecell{\textbf{$-1.3$}\\{\scriptsize$[-2.4, -0.5]$}} & \makecell{\textbf{$-0.9$}\\{\scriptsize$[-1.7, -0.1]$}} & \makecell{\textbf{$-1.7$}\\{\scriptsize$[-3.1, -0.5]$}} & \makecell{$-0.17$\\{\scriptsize$[-0.40, 0.16]$}} & \makecell{$-0.3$\\{\scriptsize$[-1.3, 0.9]$}} & \makecell{\textbf{$+0.42$}\\{\scriptsize$[0.18, 0.69]$}} & \makecell{$+0.42$\\{\scriptsize$[-0.07, 0.91]$}} & \makecell{$+0.06$\\{\scriptsize$[-0.20, 0.21]$}} \\
    shared $-$ frozen, +10 epochs & \makecell{$0.0$\\{\scriptsize$[-0.5, 0.5]$}} & \makecell{$+0.1$\\{\scriptsize$[-0.3, 0.8]$}} & \makecell{$+0.1$\\{\scriptsize$[-0.2, 0.5]$}} & \makecell{$+0.10$\\{\scriptsize$[-0.12, 0.29]$}} & \makecell{$+0.3$\\{\scriptsize$[-0.4, 1.1]$}} & \makecell{$+0.03$\\{\scriptsize$[-0.19, 0.34]$}} & \makecell{$+0.05$\\{\scriptsize$[-0.20, 0.37]$}} & \makecell{\textbf{$+0.23$}\\{\scriptsize$[0.08, 0.41]$}} \\
    per head $-$ frozen, +10 epochs & \makecell{\textbf{$+1.3$}\\{\scriptsize$[0.6, 2.3]$}} & \makecell{\textbf{$+1.0$}\\{\scriptsize$[0.4, 1.7]$}} & \makecell{\textbf{$+1.8$}\\{\scriptsize$[0.7, 3.2]$}} & \makecell{$+0.09$\\{\scriptsize$[-0.14, 0.33]$}} & \makecell{$-0.1$\\{\scriptsize$[-1.1, 0.8]$}} & \makecell{\textbf{$-0.39$}\\{\scriptsize$[-0.70, -0.03]$}} & \makecell{$-0.37$\\{\scriptsize$[-0.95, 0.22]$}} & \makecell{$+0.17$\\{\scriptsize$[-0.03, 0.50]$}} \\
    \bottomrule
  \end{tabular}
\end{table*}

\paragraph{Encoder fine-tuning (\cref{tab:ablation}(c)).}
We evaluate fine-tuning with a shared encoder or separate encoders for each head group.
Five runs start from the trained heads of \net and continue for 10 epochs on the same clip stream: 387 steps of 32 clips each, sampled with the pair weights of the training recipe (\cref{supp:C2}). Their epoch 0 corresponds to \net before adaptation.
The control keeps V-JEPA~2.1 frozen. The shared run adapts one encoder for all heads, while the per-head runs adapt separate encoders for the Semantic, Temporal, and Spatial Heads. Within the Spatial Head, stage~1, zoom, and contact decoders share their encoder; the refiner uses native crops and retains its trained weights.
The encoder follows the ViT-L fine-tuning recipe of VideoMAE~\cite{tong2022videomae}: AdamW at a learning rate of $2.5{\times}10^{-4}$ for 32 clips, layer decay 0.75, drop path 0.1, weight decay 0.05, warmup over the first 5/35 of the steps, and cosine decay.
Each head retains the optimizer, weight decay, gradient clipping, and learning rate of its recipe (\cref{tab:supp-C-schedule}), except that the Semantic and Temporal learning rates are halved for the batch of 32. The Semantic Head optimizes tool and action losses jointly, and all weights use a moving average with decay 0.999.
To account for the different loss scales, each head's gradient on the tokens is divided by its initial training loss, by the number of heads in its group, and by the number of groups. This applies equal group weights after normalization by the initial loss scale.
Training runs in bf16, and validation and test tokens run in fp32.
Each head is validated after every epoch using its loss or, for the Temporal Head, tIoU.
The shared run uses one validation-selected epoch for all head groups. The other runs select the best epoch separately for each head group.
Before adaptation, each run reproduces the validation results of \net within 0.1\%. The epoch-0 adaptation evaluation reports a tIoU of 0.591 and an R@0.7 of 40.5\%. The separate main benchmark evaluation in \cref{tab:results} reports 0.592 and 40.7\%, respectively.
\Cref{tab:supp-ft} reports the metrics and paired differences obtained by resampling test videos (\cref{supp:D2}).
Per-head fine-tuning improves classification and tool-point localization, while shared fine-tuning produces smaller changes relative to the frozen control. Selecting the best shared-model epoch for each head group still leaves it behind the separately adapted encoders on these tasks.

\subsection{Silver-style re-annotation on the Gold test set}
\label{supp:D8}

We illustrate Silver-style re-annotation by comparing propagated model-generated tracks with Gold.
\Cref{tab:results} evaluates direct How and Where predictions before propagation on the Gold test set (\cref{supp:B7}). \Cref{fig:reannot} illustrates the complete annotation procedure using Gold-only \net on three random test clips.
Tool-point errors on these examples range from 12.7 to 37.0 pixels over the 16 sampled frames, compared with the aggregate 19.15 pixels for \net in \cref{tab:results}.
The largest example-level error occurs for grasper grasp, consistent with the pair-level pattern in \cref{fig:pairsource}.

\begin{figure*}[!tp]
  \centering
  \includegraphics[width=\textwidth]{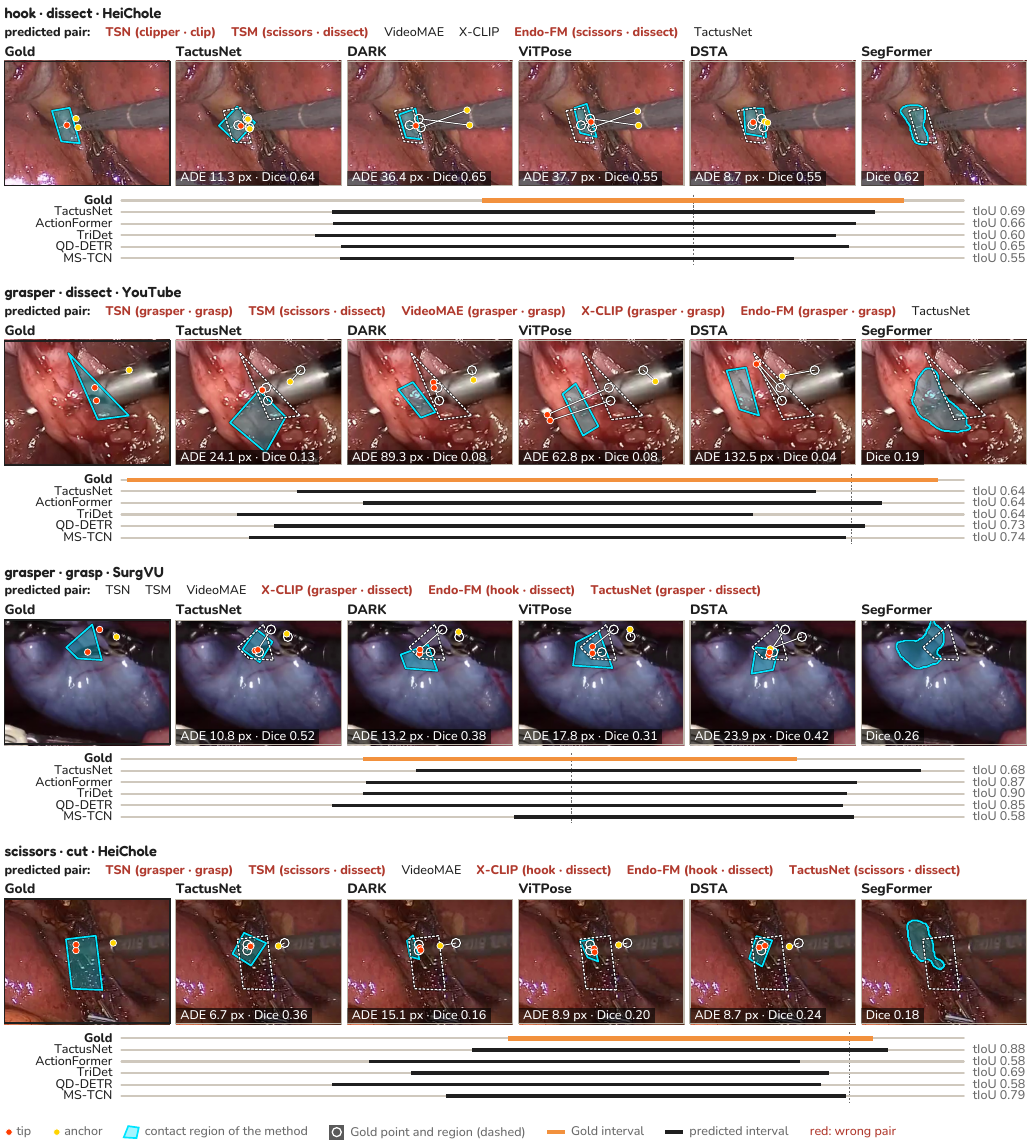}
  \caption{\textbf{Qualitative comparison of frozen \net and task-specific baselines on four random test clips.} The leading method varies across questions and clips.
  Each panel shows a sampled frame within the interaction window, comparing predicted tool points and contact regions (filled dots, cyan) with Gold (white rings, dashed outline). Labels report ADE over the clip's 16 sampled frames and Dice over its scored frames.
  Bars: the interaction window of Gold and of each When method (dashed line: the shown frame).
  Header: each classifier, with its predicted pair in red when wrong.
  As in \cref{tab:results}, the When, How, and Where methods receive the true pair.
  The clips are drawn at random with a fixed seed, at most one per pair and, where possible, one per source.}
  \label{fig:vis-baselines}
\end{figure*}

\subsection{Variation in clip-level predictions}
\label{supp:D9}

Clip-level visualizations reveal method-specific prediction patterns.
\Cref{fig:vis-baselines} compares frozen \net and the task-specific baselines on four random test clips from four different action pairs.
\net has the lowest tool-point error on three of these clips (6.7 to 24.1 pixels), consistent with its aggregate advantage in How. The keypoint baselines drift away from the tool in the YouTube example, with errors of 62.8 to 132.5 pixels.
For the other questions, example-level rankings differ more often from the dataset averages.
The four examples illustrate variation across clips and questions. \Cref{tab:results} summarizes performance on the Gold test set.

\FloatBarrier
\section{Annotation and evaluation protocols}
\label{supp:E}

Consistent annotation requires explicit temporal, spatial, and review rules.
\Cref{supp:E1,supp:E2,supp:E3} specify the Gold and Silver protocols and the tool keypoint manual.
\Cref{supp:E4} defines the corresponding evaluation protocol.

\subsection{Gold protocol: key rules}
\label{supp:E1}

\begin{figure*}[tp]
  \centering
  \includegraphics[width=\textwidth]{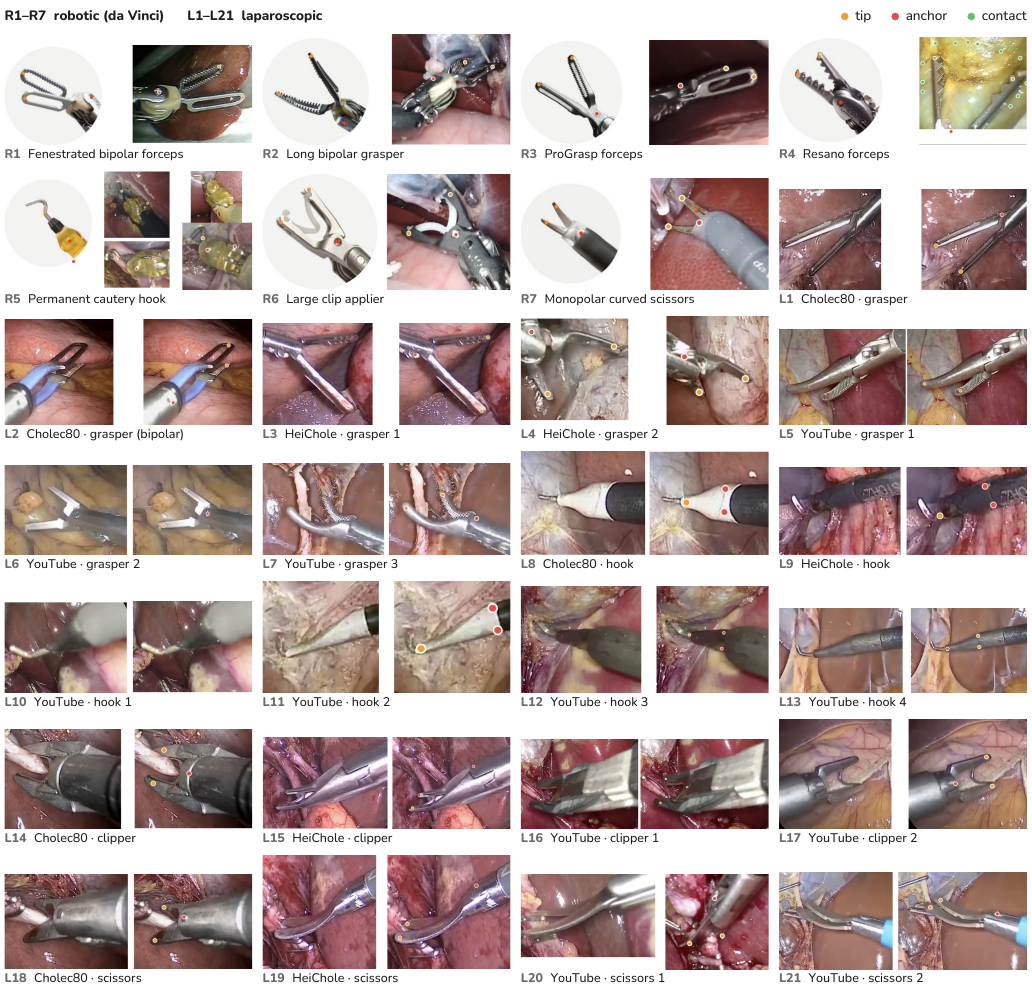}
  \caption{\textbf{Tool-specific keypoint placement illustrated by 28 robotic and laparoscopic examples.}
  R1--R7 pair product images from Intuitive's da Vinci X/Xi instrument catalog (left) with frames showing the same instrument (right).
  L1--L21 show laparoscopic examples, usually a raw frame (left) and its annotated counterpart (right).
  Archived interface examples show orange tips, red anchors, and green contact markers. The current protocol uses four tissue tracks.
  L13 reverses the tip and anchor colors, and the R4 anchor lies below the frame.}
  \label{fig:supp-E-catalog}
\end{figure*}

The general definitions in \cref{sec:definition} are operationalized by tool-specific rules for the action-window boundaries and the placement of contact points around the touched tissue (\cref{tab:supp-E-gold}).
Pauses and temporary loss of visibility remain within the same attempt. A new clip begins only when the action restarts.
The final row of \cref{tab:supp-E-gold} adds rules shared by all tools: handling unknown classes, correcting contact points that follow the tool instead of tissue, and checking visibility.
These rules are complemented by the tier-specific tolerances in \cref{supp:B3}, review procedures in \cref{supp:B4}, and shared tool-keypoint definitions illustrated in \cref{fig:supp-E-catalog} (\cref{supp:E3}).

\begin{table}[!htb]
  \centering
  \caption{\textbf{Tool-specific action-window boundaries and contact-point rules in the Gold protocol.}
  Action windows start as the tool approaches its target; the end events and contact-point placement depend on the tool. Each row specifies the usual action and target, followed by rules shared across tools.
  \Cref{supp:A2} defines the interaction-window boundaries for all six pairs.}
  \label{tab:supp-E-gold}
  \footnotesize
  \setlength{\tabcolsep}{3pt}
  \begin{tabular}{@{}>{\raggedright\arraybackslash}p{1.55cm} >{\raggedright\arraybackslash}p{2.75cm} >{\raggedright\arraybackslash}p{3.35cm}@{}}
    \toprule
    Tool (action; target) & Action window & Contact points \\
    \midrule
    Hook (dissect; tissue) & Starts as the tool moves toward the target; ends as it moves away or stops & 4 points enclosing the dissected region \\
    \addlinespace[2pt]
    Grasper (grasp; tissue) & Starts as the tool moves toward the target, usually after the jaws open; ends as it stops after retracting & 4 tissue points enclosing the grasped region with a buffer of about one tip width. Initial prompts are placed on a frame near the interaction window (\cref{supp:B2}). \\
    \addlinespace[2pt]
    Clipper (clip; duct or vessel) & Starts as the tool moves toward the target; ends as it retracts & Exactly 4 points enclosing the clipped region, on one duct or vessel \\
    \addlinespace[2pt]
    Scissors (cut; duct or vessel) & As for the clipper & 4 points enclosing the cut region, on one duct or vessel \\
    \midrule
    All tools & \multicolumn{2}{>{\raggedright\arraybackslash}p{6.4cm}@{}}{An unknown tool or action is not labeled but marked missing and reported.
    Review returns contact points that move with the tool instead of staying on tissue.
    It checks visibility only when a tool point is hidden for most of the action and returns hidden points marked visible.} \\
    \bottomrule
  \end{tabular}
\end{table}

\subsection{Silver protocol}
\label{supp:E2}

Gold and Silver share the tool categories, action definitions, and tool-keypoint definitions introduced in \cref{sec:definition}.
\Cref{fig:supp-E-catalog} illustrates these shared keypoint definitions with 28 robotic and laparoscopic examples.
For Silver, annotators provide tool, action, and temporal labels, while the generator supplies spatial prompts (\cref{supp:B6}).
\Cref{tab:supp-E-silver} summarizes the semantic and temporal annotation rules, and \cref{supp:B3} specifies the corresponding tolerances.

\begin{table}[!htb]
  \centering
  \caption{\textbf{Semantic and temporal annotation rules for Silver.} Human annotators provide the classes and times; spatial prompts are generated by the model.}
  \label{tab:supp-E-silver}
  \footnotesize
  \setlength{\tabcolsep}{4pt}
  \begin{tabular}{@{}>{\raggedright\arraybackslash}p{1.45cm} >{\raggedright\arraybackslash}p{6.4cm}@{}}
    \toprule
    Item & Rule \\
    \midrule
    Human labels & Tool, action, and times; the Silver generator places the points (\cref{supp:B6}) \\
    Action window & Covers one attempt. It starts as the tool moves toward its target with clear intent and ends when the attempt ends and the tool stops or leaves \\
    Tool classes & Grasper, hook, scissors, clipper; appearance examples are shown in \cref{fig:supp-E-catalog}. \\
    Events & Label every valid event; never merge independent events; mark unknown cases missing \\
    Priority & Label completely and consistently; quality first \\
    Workflow & Practice (demo video, protocol, 20 to 30 items; optional for experienced annotators), annotation, review of missed events, times, tool, and action by a core-team member, small fixes by the reviewer, final check \\
    \bottomrule
  \end{tabular}
\end{table}

\subsection{Tool keypoint manual}
\label{supp:E3}

Visible tool points lie at the distal, working end of the instrument, inside its outline rather than on its edge. For points that are never visible, annotators record a rough location and mark them invisible.
Within this common constraint, tip and anchor locations are defined as follows:
\begin{itemize}
  \item \textbf{Hook:} the tip is placed where the metal hook emerges from the instrument head, rather than at its far end; the two anchors lie on opposite sides of the head--shaft junction.
  \item \textbf{Grasper:} each tip is placed at the center of a jaw tip, and the anchor at the jaw pivot.
  \begin{samepage}
  \item \textbf{Scissors:} each tip is placed just inside a blade tip, and the anchor exactly at the pivot.
  \par
  \end{samepage}
  \item \textbf{Clipper:} each tip is placed just inside a jaw tip, and the anchor at the jaw base, slightly inward along the midline of the shaft end.
\end{itemize}
\Cref{fig:supp-E-catalog} applies these definitions to every example in the manual.
In \tactus, the bipolar instruments shown in R1, R2, and L2 are labeled as graspers and follow the grasper keypoint rules.

\subsection{Evaluation protocol}
\label{supp:E4}

The benchmark fixes model inputs and checkpoint selection and uses the Gold test set in \cref{supp:D1}.
Video counts follow the source video IDs. IDs associated with the same recording remain together, with the CRCD ID pair assigned to training and the two SurgVU ID pairs assigned to Silver.
Which, What, and Pair models receive only video. The annotated interaction window defines temporal and tip-derived training targets (\cref{supp:C1,supp:D2}). No benchmark model receives it as input.
Checkpoint selection uses only validation data, following \cref{supp:C2,supp:D5,supp:D7}.

\section{Release and licenses}
\label{supp:F}

The release specification connects the annotation and evaluation protocols to the distributed files, documenting their formats and source-specific license terms.

\subsection{Release format}
\label{supp:F1}

The project page is at {\urlstyle{same}\url{https://ruifreddie.github.io/TACTUS/}}. TACTUS annotations and video clips are available at {\urlstyle{same}\url{https://huggingface.co/datasets/hding2455/surgical-action-affordance}}. The source-specific licenses in \cref{supp:F2} apply to the clips. The benchmark code is available on \href{https://github.com/Ruifreddie/TACTUS}{GitHub}, and the \href{https://huggingface.co/abccbd/TACTUS-TactusNet}{model checkpoints} are hosted on Hugging Face.

\begin{table}[htbp]
  \centering
  \caption{\textbf{Release contents and their annotation fields.} The release includes action-window video clips and their TACTUS annotations from all five sources.}
  \label{tab:suppF-release}
  \footnotesize
  \setlength{\tabcolsep}{3pt}
  \begin{tabular}{@{}l >{\raggedright\arraybackslash}p{0.68\linewidth}@{}}
    \toprule
    Part & Content \\
    \midrule
    Clip index & One record per clip: video, start and end frame of the action window, tool, and action \\
    Video clips & Source-video frames spanning the annotated action window \\
    Tracks & Tool and action labels, action and interaction windows in frames, and per frame the coordinates, type (tool tip, anchor, or contact), and visibility of each point \\
    Prompt points & Key-frame points that prompted the tracker \\
    Review status & Review outcome of each record \\
    Video table & Source, clip counts, frames, frame rate, duration, and resolution per video \\
    Evaluation & Metric code (\cref{supp:D2}) and video-level Gold split assignments (\cref{supp:D1}) \\
    \net & Benchmark code and released checkpoints \\
    \bottomrule
  \end{tabular}
\end{table}

\Cref{tab:suppF-release} lists the release components. The benchmark experiments use the data partitions in \cref{supp:D1}.

\subsection{License and release of each source}
\label{supp:F2}

\begin{table}[htbp]
  \centering
  \caption{\textbf{Source licenses and the materials included in our release.}
  License entries reproduce the terms stated by each source and identify where those terms are specified.
  Annotations comprise the records, tracks, prompt points, and review status listed in \cref{tab:suppF-release}.}
  \label{tab:suppF-licenses}
  \footnotesize
  \setlength{\tabcolsep}{3pt}
  \begin{tabular}{@{}l >{\raggedright\arraybackslash}p{0.41\linewidth} >{\raggedright\arraybackslash}p{0.29\linewidth}@{}}
    \toprule
    Source & License & Released \\
    \midrule
    Cholec80~\cite{twinanda2017endonet} & CC BY-NC-SA 4.0 (dataset web page) & Clips, annotations \\
    HeiChole~\cite{wagner2023comparative} & CC BY-NC-SA, version not stated (User Notes of the paper) & Clips, annotations \\
    CRCD~\cite{oh2024comprehensive,oh2025expanded} & CC BY 4.0 (license file of the data repository) & Clips, annotations \\
    SurgVU~\cite{zia2025surgical} & CC BY-NC-SA 4.0 (official dataset repository) & Clips, annotations \\
    YouTube & Terms set by each video's rights holder & Clips, annotations \\
    \bottomrule
  \end{tabular}
\end{table}

\begin{samepage}
We distribute clips from all five sources together with their TACTUS annotations. The source-specific terms in \cref{tab:suppF-licenses} apply to the video clips.
Cholec80, HeiChole, CRCD, and SurgVU use Creative Commons licenses requiring attribution. Cholec80, HeiChole, and SurgVU also require non-commercial use and share-alike terms for adaptations.
HeiChole also requires citation of its paper~\cite{wagner2023comparative}.
Use of each YouTube video is governed by its rights holder's terms.
\end{samepage}

\clearpage
\pdfbookmark[0]{References}{shared-references}
{
  \small
  \bibliographystyle{ieeenat_fullname}
  \bibliography{refs}
}
\end{document}